\documentclass[12pt, hidelinks]{interact}

\usepackage{epstopdf}

\usepackage[square,authoryear]{natbib}
\bibpunct{[}{]}{,}{a}{}{,}

\usepackage[toc,page]{appendix}
\usepackage[pagebackref,colorlinks=true,urlcolor=blue,linkcolor=blue,citecolor=blue]{hyperref}
\usepackage[utf8]{inputenc} 
\usepackage[T1]{fontenc}    
\usepackage[american]{babel}
\usepackage{amsmath,amsfonts,amssymb,amsthm, mathtools}
\mathtoolsset{showonlyrefs}  
\usepackage{mathrsfs}
\usepackage{bm}
\usepackage{color}
\usepackage{multirow}
\usepackage{algorithm, algpseudocode, algorithmicx}
\usepackage{graphicx,wrapfig}
\usepackage{wasysym}
\usepackage{verbatim}
\usepackage{empheq}
\usepackage{grffile}
\usepackage[shortlabels]{enumitem}
\usepackage{booktabs}       
\usepackage{caption}

\usepackage{styles/general}
\usepackage{styles/paper}

\begin{document}


\title{Multivariate Bayesian Last Layer for Regression with Uncertainty Quantification and Decomposition}

\author{
  \name{Han WANG\textsuperscript{1@}, Eiji KAWASAKI\textsuperscript{1$\simeq$}, Guillaume DAMBLIN\textsuperscript{2$\simeq$}, Geoffrey DANIEL\textsuperscript{2$\simeq$}}
  \affil{\textsuperscript{1} Université Paris Saclay, CEA LIST, F-91120 Palaiseau Cedex, France}
  \affil{\textsuperscript{2} Université Paris Saclay, CEA, Service de Génie Logiciel pour la Simulation, 91191 Gif-sur-Yvette, France}
  \affil{\textsuperscript{@} Corresponding author: \href{mailto:<han.wang@cea.fr>?Subject=Your MBLL paper}{han.wang@cea.fr}, \textsuperscript{$\simeq$} Equal contribution}
}

\maketitle

\begin{abstract}
  We present new Bayesian Last Layer neural network models in the setting of multivariate regression under heteroscedastic noise, and propose EM algorithms for parameter learning. Bayesian modeling of a neural network's final layer has the attractive property of uncertainty quantification with a single forward pass. The proposed framework is capable of disentangling the aleatoric and epistemic uncertainty, and can be used to enhance a canonically trained deep neural network with uncertainty-aware capabilities.
\end{abstract}

\begin{keywords}
  Bayesian Last Layer, Multivariate regression, Uncertainty quantification, Heteroscedasticity, EM algorithm, Transfer learning.
\end{keywords}

Code and notebooks available at: \url{https://github.com/yanncalec/rundle}

\tableofcontents


\section{Introduction}\label{sec:intro}

We consider in this paper a multivariate regression model relating an input $x\in\R^\dmx$ to an output $y\in\R^\dmy$:
\begin{align}
  \label{eq:blr_linear_model_vec}
  y = A\phi(x) + \varepsilon(x)
\end{align}
and study the uncertainty quantification (UQ) in a modern setting of deep neural networks (DNNs). Here $A\sim \dmy \times \dmt$ is the regression coefficient matrix, $\phi:\R^\dmx \rightarrow \R^\dmt$ is a multivariate nonlinear feature mapping, and the heteroscedastic noise $\varepsilon(x) = \sigma(x)\varepsilon$ scales a baseline Gaussian white noise vector $\varepsilon \sim \MvNormal{0}{V}$ \footnote{We use the subscript ``vec'' for a vector-variate distribution, in order to distinguish from the matrix-variate distribution denoted without subscript in this paper. The symbol ``$\sim$'' designates either the dimension of a matrix or the probability distribution of a random variable, depending on the context.} by a scalar mapping $\sigma(x)>0$. The covariance $V\sim \dmy \times \dmy$ is either known or considered a nuisance parameter.


Multivariate regression has proven particularly useful in domains where multiple measurements are intrinsically related, such as in econometrics \citep{koop_bayesian_2010}, 
neuroscience \citep{friston_statistical_1994}, medical imaging \citep{gareen_primer_2003} and multi-task machine learning \citep{evgeniou_regularized_2004}. It provides more robust inference and more performant predictive than independent scalar regressions by modeling dependencies among all dimensions, either through correlated noise or shared latent structures across outputs.

Historically, the multivariate regression problem has been extensively studied in the literature, but mostly under the homoscedastic assumption and in moderate-dimensional settings, with fixed bases such as polynomials and splines for $\phi$, see for instance \citep{green_nonparametric_1993, hastie_generalized_1992, friedman_multivariate_1991}. In modern approaches with high dimensional inputs (such as images or multi-sensor signals) and large datasets, $\phi$ and $\sigma$ are often parameterized by learnable representations. The mapping $x \mapsto A\phi(x)$ then can be interpreted as a DNN with a linear last layer parameterized by $A$.

Building on this unifying view between classical regression and modern DNN, Bayesian Last Layer (BLL) models \citep{kristiadi_being_2020} introduce a Bayesian treatment of the final layer, while keeping the preceding layers deterministic. This hybrid approach has been proven useful for UQ featuring principled training and efficient Bayesian inference. We introduce in this work a novel BLL framework for multivariate nonlinear regression under heteroscedastic noise, contributing to single-pass uncertainty decomposition and quantification in multidimensional settings.

\subsection{Backgrounds and Related Works}
\subsubsection*{Uncertainty Quantification}
Recent advances in DNNs have spurred significant interest in uncertainty quantification for predictive models \citep{gawlikowski_survey_2023}. DNN predictions are subject to two key types of uncertainty: aleatoric and epistemic. The former stems from inherent randomness and measurement noise or lack of information due to incomplete observations, while the later arises from model limitations, such as architecture choice and finite data. Ensuring the safe deployment of DNNs hinges on providing trustworthy predictions, often represented by prediction intervals that account for both uncertainties. Various UQ methods have been proposed, including Bayesian Neural Networks (BNNs) \citep{wilson_bayesian_2020}, Monte Carlo Dropout \citep{gal_dropout_2016}, and Ensembles \citep{lakshminarayanan_simple_2017}. More recently, hybrid approaches combining Bayesian principles with single and deterministic DNNs have emerged, offering computational efficiency while maintaining uncertainty estimation. Among these, Deep Evidential Regression (DER) \citep{amini_deep_2020} provides closed-form uncertainty decomposition by parameterizing the predictive distribution. Further extensions include multivariate DER \citep{meinert_multivariate_2021}, along with analyses of parameter identifiability and model effectiveness \citep{meinert_unreasonable_2023}.

\subsubsection*{Bayesian Last Layer Models}
Bayesian Last Layer generalizes Bayesian Linear Regression (BLR) by interpreting the penultimate layer’s activations as nonlinear feature mappings, and can be viewed as a restricted BNN \citep{sharma_bayesian_2023}. The central idea is to retain the computational efficiency, interpretability, and analytical tractability of classical Bayesian models by confining stochasticity to the last layer and pushing all nonlinearities into deterministic feature mappings. A Normal distribution-based BLL can also be interpreted as a Gaussian Process with a feature-transformed linear kernel, which naturally connects BLL to Deep Kernel Learning (DKL) \citep{wilson_deep_2016}. Initial formulations of BLL \citep{lazaro-gredilla_marginalized_2010} were later applied to Bayesian optimization \citep{snoek_scalable_2015} and benchmarked against joint feature-parameter learning techniques \citep{ober_benchmarking_2019}. Recent developments address the overconfidence in out-of-distribution predictions \citep{watson_latent_2021}, extend the framework to multivariate outputs \citep{fiedler_improved_2023}, classification problems \citep{harrison_variational_2024}, and the heteroscedastic noise setting \citep{harrison_heteroscedastic_2025}. Sampling-based implicit priors on the last layer weights have also been explored to further increase the expressive power of BLL models \citep{xu_flexible_2024}.

\subsubsection*{Training via VI}
Mini-batch stochastic gradient descent (SGD) is the standard tool for training DNNs on large datasets. However, for a Bayesian model like the BLL, marginal likelihood is not a suitable objective function for SGD due to biased gradient estimate, as mentioned in precedent works \citep{harrison_variational_2024, watson_latent_2021}. In variational BLL (VBLL) \citep{harrison_variational_2024}, the authors used ELBO-based variational inference (VI), which jointly trains the DNN component and estimates the parameters of a variational posterior. This method enjoys a mini-batch-compatible objective function and requires the prior to be specified.

\subsubsection*{Hyperparameter estimation}
As an empirical Bayes method, the \emph{evidential framework} \citep{mackay_bayesian_1992} is a principled way for estimating \emph{hyperparameters} (\ie prior parameters) in regression-related UQ. Also known as the \emph{type-II} MLE, it selects optimal hyperparameters by maximizing the \emph{evidence}, or marginal likelihood function of the data, with repect to the prior. Precedent works claimed this principle ensures the appropriate level of model complexity to avoid underfitting or overfitting\footnote{Actually, our results show that the overfitting can still occur since the joint optimization of the unregularized evidence leads to the MLE solution.} due to a misspecified prior (\ie too strong or too weak). To our knowledge, the issue of hyperparameter estimation has not yet been addressed in BLL literature so far. For example, in \citep{blundell_weight_2015, harrison_variational_2024}, the prior was specified manually and used uniquely as initialization to obtain the variational posterior, which can be clearly affected by a poor choice of hyperparameters.

\subsection{Contributions}
Our new BLL regression models aim at improving existing ones by extending them to multivariate and heteroscedastic situations, and by making them suited for pratical scenarios like Transfer Learning, where pre-trained DNNs can be enhanced with UQ capabilities for new domains. We leverage matrix-variate distributions, which provide compact representations and open paths toward efficient optimization algorithms.

Our models are built on the classical Bayesian linear regression, reviewed and extended in Section~\ref{sec:bayesian_regression}, which provides closed-form decompositions for aleatoric and epistemic uncertainty. Unlike the classical regression with a fixed basis, a BLL model contains a learnable feature mapping $\phi$ and a noise scaling $\sigma$, both parameterized by DNN. Together with prior parameters, these are the variables of a BLL model that need to be fixed by the user. The closed-form posterior then provides sampling-free and single-pass predictions with UQ.

We address the issues of hyperparameter estimation and training of DNN in a unified evidential framework. Contrary to the common belief that all BLL parameters can be jointly learned by maximizing marginal likelihood, we analyze the evidence objective function in Section~\ref{sec:blr_parm_estim} and prove that joint optimization of BLL parameters leads to a degenerate MLE, and propose stabilizing the pathological proof function by keeping the mean fixed or introducing a regularizing hyperprior on the covariance. Our BLL model relies on the important modeling assumption of \emph{shared covariance} between the baseline noise and the prior. In the more general setting of unrelated covariances, we lose the compact matrix-variate representation of the posterior distribution. In Section~\ref{sec:var_post} we discuss the approach of VBLL which is an ELBO-based VI without assumption of shared covariance, and point out its connection with the EM algorithm.

Training of DNN weights in a BLL model becomes more challenging, since the gradient estimate on mini-batches is no longer guaranteed unbiased, due to the loss of sample independence in the evidence function. We propose in Section~\ref{sec:elbo-em} a numerically efficient EM algorithm which is compatible with modern DNN training pipelines, and admits closed-form update formulae for hyperparameters. It cleanly separates deterministic DNN learning from hyperparameters inference, permitting in this way the application in the transfer learning scenario. Notably, the EM algorithm can be viewed as iterative application of the ELBO-based VI. Its performance is demonstrated through numerical experiements.

Finally in Section~\ref{sec:mvblr_T}, the theoretical and algorithmic framework is extended to the general case where the noise covariance is unknown via the matrix-T distribution, followed by a discussion of limitations and future work.

For clarity of presentation, proofs of main results are relegated to Appendix~\ref{sec:proofs_tech}.

\section{Bayesian Last Layer Model and Parameter Estimation}
\label{sec:mvblr}

Suppose that $N$ independent observations $\set{(x_i, y_i)}_{i=1:N}$ have been realized
with underlying i.i.d. noise ${\noise_i}$. We define the following matrices
\begin{align}
  Y \defeq [-y_i-], \ \ \Phi \defeq [-\phi(x_i)-], \ \ \Noise \defeq \bracket{-\sigma(x_i)\varepsilon_i-}
\end{align}
as horizontal concatenation of column vectors, which have dimensions $\dmy\times\dmN$, $\dmt\times\dmN$ and $\dmy\times\dmN$ respectively. The regression model \eqref{eq:blr_linear_model_vec} expressed in matrix form becomes
\begin{align}
  \label{eq:blr_linear_model}
  Y = A\Phi + \Noise
\end{align}
The classical Bayesian linear regression (BLR) model for \eqref{eq:blr_linear_model} provides an analytically tractable framework for inference through conjugate priors. A large body of classical text exists on BLR, see for example \citep{bishop_bayesian_2003, box_bayesian_2011, murphy_machine_2012, minka_bayesian_2010}. BLR enables exact posterior computation and closed-form quantification of both aleatoric and epistemic uncertainties, offering significant computational advantages over sampling-based methods. Analyses and notations in the next section are mainly adapted from \citep{minka_bayesian_2010}.

In the following we denote by $\Var{\cdot}$ the row covariance operator $\Var{Y}\defeq\Exp{\paren{Y-\Exp{Y}}\trpp{Y-\Exp{Y}}}$. The column-stacked vector of a matrix is denoted $\vectorize{\cdot}$. We write $\exptr(\cdot)$ for $\exp(\trace(\cdot))$ and $\abs{\cdot}$ for the determinant. The symbol $\otimes$ stands for Kronecker product of matrices or tensor product of distributions, depending on the context. Important symbols and their definitions are summarized in Table~\ref{tab:symbols}. Other notations will be introduced as we go along.

\begin{table}[H]
  \centering
  \caption{Symbols used in Section~\ref{sec:bayesian_regression} (matrix-Normal model) and Section~\ref{sec:bayesian_regression_T} (matrix-T model).}
  \label{tab:symbols}
  \begin{tabular}{cc}
    \toprule 
    \bfseries Symbol & \bfseries Definition\\
    \midrule 
    $\Phi, Y, X$ & data matrices\\
    $A$ & regression coefficients\\
    $M, K, V$ & prior parameters\\
    $D$ & $\diag\bracket{\sigma^2(x_i)}$\\
    $\Pxx$ & $D + \Phit K \Phi$\\
    $\Syx$ & $Y\inv D \Phit + M\inv K$\\
    $\Sxx$ & $\inv K + \Phi \inv D \Phit$ \\
    \midrule 
    $\Spr$ & prior parameter\\
    $\Syy$ & $Y \inv D Y^\top+ MK^{-1} M^\top$ \\
    $\Sycx$  & $\Syy - \Syx\iSxx\Syx^\top$ \\
    $\Sycxa$ & $\Syy+A\Sxx A^\top - \paren{\Syx A^\top + A\Syx^\top}$\\
    \midrule
    $\theta$ & BLL hyperparameters \\
    $\wt$  & weights of DNN $\phi, \sigma$ \\
    $\xi$  & $\theta \cup \wt$ \\
    \bottomrule 
  \end{tabular}
\end{table}

\subsection{Bayesian Regression}
\label{sec:bayesian_regression}

Following standard BLR theory, we specify a matrix-Normal \footnote{Various equivalent definitions for matrix-variate distributions exist in the literature, and we adopt the conventions from \citep{gupta_matrix_1999}.} prior for $A$:
\begin{align}
  \label{eq:blr_prior_case1}
  A\sim \MaNormal{M}{V}{K},
\end{align}
with $M\sim \dmy\times\dmt$, $V\sim\dmy\times\dmy$ and $K\sim\dmt\times\dmt$ being prior parameters ($V$ and $K$ are symmetric and positive definite), and the density function is given by:
\begin{align}
  \MaNormal{A;M}{V}{K}
  \propto \abs{V}^{-\frac \dmt 2}\abs{K}^{-\frac \dmy 2}\exptrp{-\half \inv V \paren{A-M}\inv K \trpp{A-M}}
\end{align}
This distribution generalizes the vector-Normal distribution by introducing separate covariance matrices $V$ (for rows) and $K$ (for columns). The parameterization is clearly non-unique, as $\MaNormal{M}{cV}{c^{-1}K}$ represents the same distribution for any scalar $c>0$.
Hence $K$ cannot be uniquely identified unless $V$ is known. It should be noticed that the column vectorization of \eqref{eq:blr_prior_case1} follows a vector-Normal distribution $\vectorize{A}\sim\MvNormal{\vectorize{M}}{K\otimes V}$, but the rearrangement of a vector-Normal distribution in matrix form does not yield a matrix-Normal distribution in general. As a useful property about the matrix-Normal distribution, let us mention that for any square matrix $Q$ of appropriate dimensions it holds
\begin{align}
  \label{eq:mvd_Normal_cov}
  \Exp{\paren{A-M} Q \paren{A-M}^\top} = \trace\paren{Q^\top K} V
\end{align}
This identity will prove essential for subsequent uncertainty quantification.

Define $D\defeq \diag[\sigma^2(x_i)]$ the diagonal matrix of noise variances which are strictly positive, the likelihood of $Y$ given $A$ is
\begin{align}
  \label{eq:blr_likelihood_case1}
  Y|A\sim\MaNormal{A\Phi}{V}{D},
\end{align}
The conjugate prior then yields closed-form expressions for other related distributions, as summarized in Theorem~\ref{thm:blr_case1}. These results generalize classical BLR theory to the matrix-variate heteroscedastic setting, containing the classical case as a special instance. Notably, they also show that BLL is equivalent to a multivariate Gaussian process with the linear kernel $k(x,x') = \phi(x)^\top K \phi(x')$.

\begin{theorem}[Bayesian regression]
  \label{thm:blr_case1}
  Suppose that the noise covariance $V$ is known. With the symbols defined in Table~\ref{tab:symbols}, it holds the following distributions:
  \begin{enumerate}
    \item  The joint distribution of $[A\ Y] \sim \dmy\times (\dmt+\dmN)$ is
      \begin{align}
        \label{eq:blr_joint_case1}
        [A\ Y]\sim \MaNormal{[M\  M\Phi]}{V}{\PsiMatKX}
      \end{align}
    \item The evidence is
      \begin{align}
        \label{eq:blr_evidence_case1}
        Y\sim\MaNormal{M\Phi}{V}{\Pxx}
      \end{align}
    \item The posterior is
      \begin{align}
        \label{eq:blr_posterior_case1}
        A|Y\sim\MaNormal{\Syx\iSxx}{V}{\iSxx}
      \end{align}
    \item Given $L$ new inputs $X_*\sim \dmx\times L$, the posterior predictive is
      \begin{align}
        \label{eq:blr_predictive_case1}
        Y_* | Y \sim \MaNormal{\Syx\iSxx \Phi_*}{V}{D_*+\Phi_*^\top \iSxx \Phi_*}
      \end{align}
      where $\Phi_*\defeq\phi(X_*)$ and $D_* \defeq \diag[\sigma^2(X_*)]$ are formed similarly as $\Phi$ and $D$.
  \end{enumerate}
\end{theorem}

\begin{remark}
  With a new data point $(x_{N+1}, y_{N+1})$, the matrix $\Syx\iSxx$ and $\iSxx$ involved in the posterior \eqref{eq:blr_posterior_case1} and the predictive \eqref{eq:blr_predictive_case1} can be updated by taking their old values as the prior parameters $M$ and $K$ respectively. This is parallele to the Bayesian update formula of the classical BLR theory.
\end{remark}

\paragraph*{Helper identities}
Below are some useful identities for some symbols defined in Table~\ref{tab:symbols}, which can be established using the Woodbury matrix inversion formula.
\begin{equation}
  \label{eq:blr_helper_id_1}
  \begin{aligned}
    \iSxx &= K - K \Phi \iOmega \Phit K \\
    \iPxx &= \inv D - \inv D \Phit \iSxx \Phi \inv D \\
    \Syx\iSxx &= M+\paren{Y-M\Phi}\iPxx \Phit K,
  \end{aligned}
\end{equation}
Moreover, assuming that $\dmN\geq\max(\dmt, \dmy)$ and $\Phi$ has full rank, then
\begin{equation}
  \label{eq:blr_helper_id_2}
  \begin{aligned}
    \invp{\Phi\iPxx \Phit} & = K + \invp{\Phi \inv D \Phit} \\
    \iPxx \Phit \invp{\Phi\iPxx \Phit} &= \inv D \Phit \invp{\Phi\inv D \Phit}.
  \end{aligned}
\end{equation}

\paragraph*{Bayesian regression compared to MLE}
The classical maximum likelihood estimator (MLE) for $A$ is obtained by maximizing the log-likelihood $\ln \Pdf{Y|A}$, Under the assumption that ${\Phi\inv D\Phit}$ is invertible, the unique solution can be expressed equivalently through the weighted least square formulation (of the Frobenius norm)
\begin{align}
  \label{eq:lsq_solution_AD}
  \hat A_D = \argmin_A \half \norm{A\Phi D^{-\half}-Y D^{-\half}}^2.
\end{align}
and reads $\hat A_D=Y\inv D\Phit\invp{\Phi\inv D\Phit}$. The next result shows that the variance of Bayesian posterior predictive is always smaller than the least square one.

\begin{proposition}\label{prop:bayes_leastsq_comparison}
  Assume $\dmN\geq\max(\dmt, \dmy)$ and $\Phi$ has full rank. For a prediction point $x_*$, let $\hat y_* = \SyxiSxx \phi(x_*)$ and $\hat y_D = \hat A_D \phi(x_*)$. Then it holds
  \begin{align}
    \tracep{\Var{\hat y_D} - \Var{\hat y_*}} = {\phi(x_*)^\top \paren{\invp{\Phi \inv D \Phit}+\iSxx} \phi(x_*)} \cdot \tracep{V} \geq 0.
  \end{align}
\end{proposition}
Since the least square solution actually also corresponds to the optimal weight of the (linear) last layer of a probability density network, this proposition indicates that the BLL model always provides better regularization than the canonical training.

\subsubsection{Quantification of uncertainties}
\label{sec:blr_UQ_case1}
Uncertainty quantification is essential for making reliable predictions and can be generally categorized into \emph{epistemic and aleatoric} uncertainty, which origins respectively from the limited knowledge of model and from the intrinsic randomness in data. One possible way to quantify them in the regression problem is through the total variance decomposition \citep{amini_deep_2020}. We extend this idea to the posterior predictive in a matrix-variate version.

\begin{proposition}[Decomposition of uncertainties]
  \label{prop:blr_UQ_case1}
  Let $\lambda=\set{A}$ and $\theta=\set{M,K}$ be respectively the likelihood and prior parameters of the Bayesian regression model. Given the data $Y$ and new points $X_*$, it holds the total variance decomposition for the posterior predictive of $Y_*$
  \begin{align}
    \label{eq:blr_total_var_decomp_predictive}
    \Var{Y_*|Y, \theta} = \underbrace{\Exp{\Var{Y_*|\lambda}|Y,\theta}}_{\text{Aleatoric}} + \underbrace{\Var{\Exp{Y_*|\lambda}|Y,\theta}}_{\text{Epistemic}}
  \end{align}
  where the aleatoric and epistemic uncertainty are explicitly given as
  \begin{subequations}
    \label{eq:blr_uncertainties_case1}
    \begin{align}
      \Exp{\Var{Y_*|\lambda}|Y,\theta} &= \trace\paren{\Ds}V, \label{eq:blr_uncertainties_case1_aleatoric} \\
      \Var{\Exp{Y_*|\lambda}|Y,\theta} &= \trace\paren{\Phit_* \iSxx \Phi_*} V. \label{eq:blr_uncertainties_case1_epistemic}
    \end{align}
  \end{subequations}
\end{proposition}

\paragraph*{Asymptotic of epistemic uncertainty}
Epistemic uncertainty, which reflects model uncertainty due to limited data and knowledge, typically increases in regions with sparse training data coverage and decays asymptotically with additional observations. To formalize this behavior, we establish the following theoretical result: Under standard regularity conditions, the epistemic uncertainty --- quantified through $\trace(\Phi_*^\top \Sxx^{-1} \Phi_*)$ --- scales as $O(1/\dmN)$ where $N$ represents the sample size. This justifies the empirical observation that epistemic uncertainty diminishes with increasing data.

\begin{proposition}
  \label{prop:asymptotic_epistemic}
  Assume a probability distribution $\opPrb_x$ for the input variable $x$, and $\phi(\cdot)$ and $\sigma(\cdot)$ fulfill
  \begin{align}
    \label{eq:PDP_condition}
    \Exp[x\sim\opPrb_x]{\frac{{\phi(x)\phi(x)^\top}}{\sigma^2(x)}} \succ 0.
  \end{align}
  Then for $\dmN$ i.i.d. samples $\set{x_i}$ from $\opPrb_x$, it holds almost surely $\iSxx=O(1/\dmN)$ as $\dmN$ increases.
\end{proposition}

\subsection{Parameter Estimation by Evidential Framework}
\label{sec:blr_parm_estim}
We assume hereafter the baseline noise covariance $V$ is known. Let $\theta=\set{M,K}$ be prior parameters and $\wt=\set{\wtp, \wts}$ be weights of DNNs. Denote the ensemble of parameters to be optimized as
\begin{align}
  \label{eq:def_xi_case1}
  \xi=\theta\cup \wt=\set{M,K,\wtp, \wts}
\end{align}
A principled way for parameter optimization of a Bayesian model is through the \emph{evidential framework} \citep{mackay_bayesian_1992}, summarized in Appendix~\ref{sec:ev_framework}. This framework looks for the optimal parameters $\xi$ by maximizing
\begin{align}
  \label{eq:LnEV}
  \ln \Pdf{\xi|Y} \simeq \ln \Pdf{Y|\xi} + \ln \pi\paren{\xi}
\end{align}
where $\pi$ is an optional hyperprior for $\xi$ providing regularization to the evidence function $\Pdf{Y|\xi}$. We study in the following the existence and uniqueness of the optimizer.

\subsubsection{Study of the evidential loss function}
\label{sec:ev_loss_case1}
For simplicity, we drop the hyperprior in \eqref{eq:LnEV}, and assume first DNN weights are fixed in $\xi$. Up to irrelevant constants, the loss function is equivalent to the negative log-evidence $-\ln\MaNormal{Y;M\Phi}{V}{\Pxx}$ and reads
\begin{align}
  \label{eq:ev_loss_case1}
  \lL(M,K) \defeq \frac{\dmy}{2}\ln\abs{\Pxx} + \half\tracep{\iPxx E^\top \inv V E} 
\end{align}
with the shorthand $E \defeq Y-M\Phi$. Partial derivatives can be obtained by applying matrix calculus \citep{petersen_matrix_2008}, or via automatic tools \citep{laue_computing_2018}. In particular, since $M$ is unconstrained, we have
\begin{equation}
  \label{eq:bll_mle_derv_case1}
  \begin{aligned}
    \pypx{\lL}{M} & = -\inv V E \iPxx \trp{\Phi}.
  \end{aligned}
\end{equation}
We look for a minimizer in $(M,K)\in\lM_{\dmy,\dmt} \times \Spp^\dmt$, \ie the product space of $\dmy\times\dmt$-dimensional matrix and $\dmt$-dimensional symmetric positive definite matrix. Using the criteria of first order optimality, we can prove that the joint optimization admits a degenerate global minimizer given by
\begin{align}
  \label{eq:bll_mle_MK_case1}
  \hat M = Y \inv D \Phit \invp{\Phi\inv D \Phit}, \ \ \hat K=0
\end{align}
as stated in the next result.

\begin{theorem}[Joint minimization]
  \label{thm:bll_mle_KVM}
  Assume that $\dmN\geq \max(\dmt,\dmy)$ and $\Phi$ has full rank. The following statements hold:
  \begin{enumerate}
    \item
      The joint minimization of the loss with respect to $(M, K)\in\lM_{\dmy\times\dmt} \times \Spp^\dmt$ is reduced to the minimization of $K\in\Spp^\dmt$ alone:
      \begin{align}
        \min_{M,K} \lL(M,K) = \min_K \lL(\hat M, K)
      \end{align}
      where the unique unbiased minimizer $\hat M$ is $K$-independent and equals to the least square solution \eqref{eq:lsq_solution_AD}.
    \item
      With $\hat M$, the loss is concave and strictly increasing in $K$, that is $\lL(\hat M, K) > \lL(\hat M, K')$ for any $K\succ K'$ in \emph{Löwner} order\footnote{For definition see Section~\ref{sec:annex_convex}.}. In particular the singular point $\hat K=0$ is the unique global minimizer.
  \end{enumerate}
\end{theorem}

Since $\hat K=0$ implies $\iSxx=0$, the epistemic uncertainty defined in \eqref{eq:blr_uncertainties_case1_epistemic} vanishes and the model becomes over-confident.
Therefore the unregularized evidence objective is pathological for joint parameter estimation of BLL in view of UQ.  The same statement holds by taking into consideration the DNN weights $\wt$ and treate $\lL$ as function of $(M, K, \wtp, \wts)$: It is an easy corollary of Theorem~\ref{thm:bll_mle_KVM} that, as long as the DNN weights fulfill the same assumption, the joint minimization of $\theta\cup \wt$ is reduced to the minimization of $\wt$ alone with the same degenerate global minimizer $\hat\theta=(\hat M, \hat K)$ as $\theta$.


\subsubsection{Stabilization of the evidential loss}
\label{sec:regularize_ev_loss}
We present two methods to stabilize the pathological evidential loss: by holding $M$ fixed or by introducing a hyperprior $\pi(\cdot)$ on $K$. Both techniques modify the loss to be the difference of two decreasing convex functions in $K$. The two methods can be combined together, and the regularized loss is an instance of problems studied in \emph{D.C. programming} \citep{le_thi_dc_2018}, for which global optimality guarantees generally do not exist.

\begin{figure}
  \centering
  \includegraphics[width=1.\textwidth]{{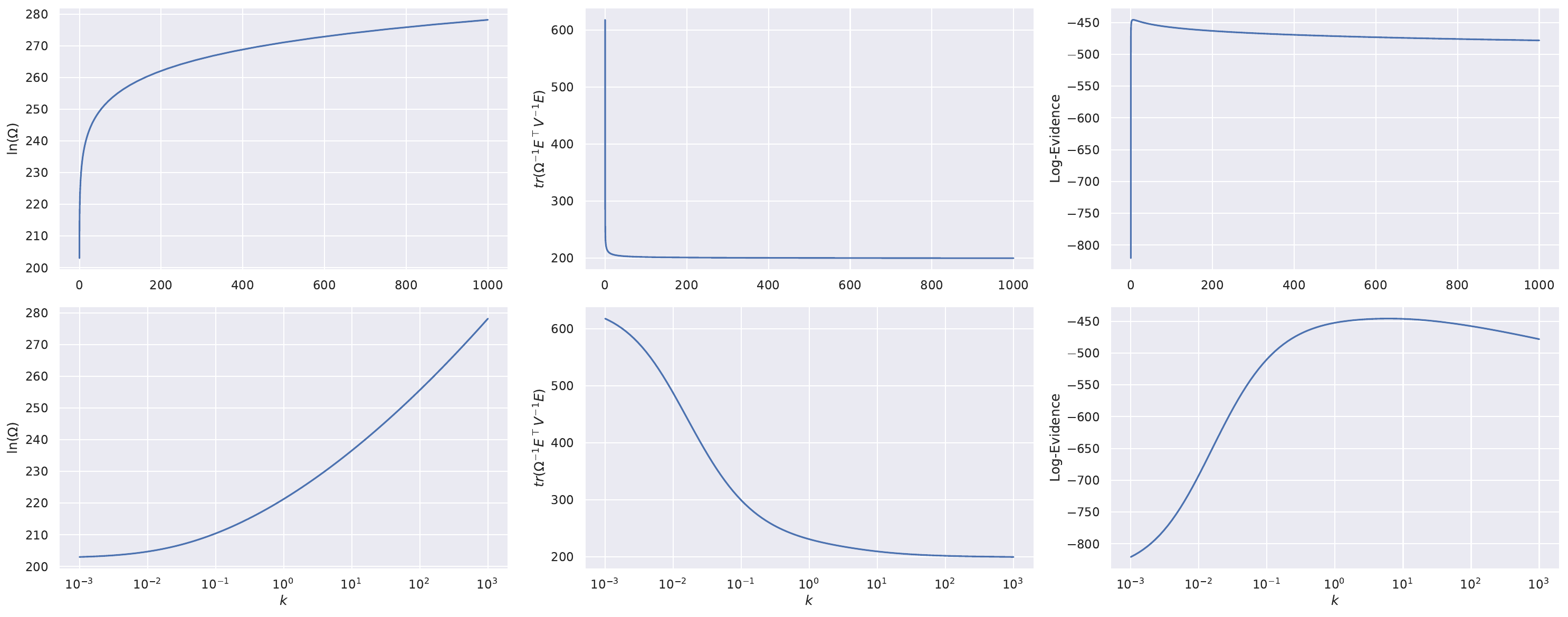}}
  \caption{Components of log-evidence $\ln\abs{\Pxx}$ and  $\tracep{\iPxx E^\top \inv V E}$ as a function of $k$ for the example in Figure~\ref{fig:em_fixedbasis}. Abscissa of pannels in the second row are in logarithmic scale.}
  \label{fig:comp_lnev_k}
\end{figure}

 \paragraph*{Fixed $M$}
Joint optimization of location and scale parameters in a Bayesian model is known to be problematic, and in precedent works \citep{mackay_bayesian_1992, tipping_sparse_2001} only the covariance $K$ is optimized by holding the mean $M=0$ fixed (which doe not seem to be a limiting factor for the predictive power, at least for 1d interpolation problems). Proposition~\ref{prop:logdet_traceinv} allows us to see the following structural properties of the loss \eqref{eq:ev_loss_case1} as function of $K$: the term $\ln\abs{\Pxx}$ is concave and monotonically increasing, and the term $\tracep{\iPxx E^\top \inv V E}$ is convex and monotonically decreasing. As an example, these two components and the log evidence of the data in Figure~\ref{fig:em_fixedbasis} are plotted in Figure~\ref{fig:comp_lnev_k} as function of the isotropic $K=kI$. It can be observed that the evidence is indeed maximized at some $k\neq 0$ in this example. However, this regularization is not sufficient to avoid the degenerate solution in general. Precedent theoretical analysis \citep{tipping_sparse_2001} showed that by fixing $M=0$, the vanished covariance $K=0$ may still emerge as a global optimum. This seemingly disadvantage in view of UQ is actually exploited to prune insignificant features (those collapsed to 0) in order to enhance sparsity of the Bayesian model, see \eg \citep{graves_practical_2011, bishop_bayesian_2003}.

\begin{remark}
Given a canonically trained DNN (using the MSE loss), one might be tempted to set $M$ equal to the weights of the last layer rather than choosing $M=0$. However, these last-layer weights correspond exactly to the least-squares estimator $\hat M$. As shown in Theorem~\ref{thm:bll_mle_KVM}, this choice inevitably leads to the degenerate solution $K=0$. To overcome this issue, we introduce in the next a hyperprior-based regularization strategy that allows arbitrary choices of $M$ or joint optimization of hyperparameters while avoiding singular solutions.
\end{remark}



\paragraph*{Regularization by hyperprior}
A second method consists in placing an Inverse Wishart hyperprior $\pi=\InvWishart{\HypK}{\NdfK}$ on $K$, which is a standard conjugate prior for covariance matrix in Bayesian modeling. The definition of Inverse Wishart distribution is given in Section~\ref{sec:bayesian_regression_T}. The loss function is then equivalent to the negative of \eqref{eq:LnEV} while still being minimized at $\hat M$. However, evaluated at $\hat M$ it becomes
\begin{align}
    \label{eq:ev_loss_case1_hypK}
  \lL(\hat M, K) \simeq \frac{\dmy}{2}\ln\abs{\Pxx} + \frac{\NdfK}{2}\ln\abs{K} + \half\tracep{\inv K \HypK}
\end{align}
which behaves as $O(\inv{\norm{K}})$ as $K\rightarrow 0$. Therefore $K=0$ is no more an optimum as for the unregularized evidential loss. The term $\ln\abs{\Pxx}$ is concave and monotonically increasing in $K$ (a fact revealed by the proof of Theorem~\ref{thm:bll_mle_KVM}), so is $\ln\abs{K}$. On the other hand, the trace term is convex and monotonically decreasing in $K$. Putting together, the regularized loss \eqref{eq:ev_loss_case1_hypK} is the difference of two decreasing convex functions.

Since the regularization effect is provided by the trace term in \eqref{eq:ev_loss_case1_hypK}, the degree of freedom $\NdfK$ can be kept small (\eg $\NdfK=1$) in practice. For the location parameter $\HypK$, one can choose a large value so that the hyperprior becomes uninformative. For the transfer learning scenario of Section~\ref{sec:em_transfer_learning} with pretrained DNNs, we can set $\HypK=\invp{\Phi\inv D\Phi^\top}$ to approximate the effect of a scale invariant prior \footnote{An inference for regression is \emph{scale invariant} if the regression matrix $\hat A$ is inferred from the data $(Y,X)$, then $B\hat A\inv C$ should be infered from the scaled data $(BY, CX)$ for some invertible matrix $B,C$. This is often a desired property and is met by the least square solution.}.

\subsection{Variational posterior}
\label{sec:var_post}

Results established so far are based on the modeling assumption that the prior \eqref{eq:blr_prior_case1} shares the same covariance $V$ with the baseline noise distribution $\varepsilon\sim\MvNormal{0}{V}$ --- a well-known BLR technique that both reduces hyperparameters and enables analytical posterior computation through conjugacy, see \eg \citep{murphy_machine_2012, minka_bayesian_2010}. Here we briefly discuss BLL models in the more general setting where the baseline noise covariance, now named $U$, is unrelated to the prior covariance $V$, and review the algorithmic connections with the variational BLL (VBLL) method \citep{harrison_variational_2024}.

In this general setting one can still work out a fully analytical Bayesian model, but at the cost of losing the compact matrix-variate representation of Theorem~\ref{thm:blr_case1}. In particular, the marginal likelihood of $Y$ can only be expressed in a vector-variate form:
\begin{align}
  \label{eq:Yvec_marginal_UV}
  \vectorize{Y} \sim \MvNormal{\vectorize{M\Phi}}{D \otimes U + \paren{\Phi^\top K \Phi}\otimes V}.
\end{align}
The posterior and the predictive distributions can also be established analytically and are given in Theorem~\ref{thm:blr_UV_case1}. Compared to the matrix-variate counterpart \eqref{eq:blr_evidence_case1}, this form is cumbersome and computationally inefficient due to its full-size covariance matrix.

One workaround then is to seeking a matrix-variate variational posterior via ELBO maximization. As a lower bound of the marginal log-likelihood, the ELBO relates the data $Y$, the variational posterior $q$ and the prior $\pi$, and is defined as
\begin{align}
  \label{eq:elbo_def}
  \ELBO{Y}{q}{\pi}\defeq \Exp[A\sim q]{\ln\Pdf{Y|A}} - \KL{q}{\pi}
\end{align}
where $\opKL$ denotes the Kullback-Leibler divergence. Let $\lN_\theta$ denote a fixed matrix-Normal prior of parameters $\theta$, we seek for the optimal variational posterior $\lN_\ttheta$ via
\begin{align}
  \label{eq:elbo_post_theta}
  \max_{\ttheta} \ELBO{Y}{\lN_{\ttheta}}{\lN_\theta}
\end{align}
This is the approach adopted by variational BLL (VBLL) in \citep{harrison_variational_2024}. This ELBO objective admits closed-form expression, which leads to coordinate-wise update for hyperparameters and unbiased gradient estimates for DNN parameters.
For shared covariance $U=V$, which is the setting of our BLL models, \eqref{eq:elbo_post_theta} just returns the true matrix-variate posterior $\Pdf{A|Y}$ of \eqref{eq:blr_posterior_case1}. Notably, it can be interpreted as one step of the EM algorithm developed in Section~\ref{sec:elbo-em}. Details are presented in Appendix~\ref{sec:equiv_vbll_em}.

\section{Optimization with EM}\label{sec:elbo-em}

Parameters of BLL model can be numerically optimized through the MAP problem $\max_\xi \ln\Pdf{\xi|Y}$, defined in \eqref{eq:LnEV}. However, due to the complex form of the evidence function, a direct SGD method applied to all BLL parameters without resorting to matrix calculation would be inefficient; moreover, unbiased estimation of the gradient on mini-batches is no longer guaranteed.

We employ here the Expectation Maximization (EM) method for the MAP problem, which treats hyperparameters and DNN weights separately in efficient ways. Originally developed for MLE with missing data, EM extends naturally to Bayesian inference by treating the likelihood parameters as latent variables \citep{lange_robust_1989}. Sampling-based EM algorithms have been proposed previously  for optimization of DNN parameters (with fixed hyperparameters) \citep{nejatbakhsh_hybridbnn_2024}. In contrast, our EM algorithm are based on close-form evaluation of E-step, allowing efficient joint optimization of all parameters.


\subsection{EM Algorithm for BLL}\label{sec:EM_algo_case1}

The E-step of EM consists in evaluating the surrogate function
\begin{align}
  \label{eq:EM_Qfunc}
  Q(\xi;\txi) \defeq \Exp[{A\sim\Pdf[\txi]{\cdot|Y}}]{\ln\Pdf[\xi]{[A, Y]}} + \ln\pi(\xi)
\end{align}
where $\txi = \tilde\theta \cup \tilde\wt = \set{\tM, \tK, \twtp, \twts}$ is the set of parameters \eqref{eq:def_xi_case1} from the last M-step. Here the posterior $\Pdf[\txi]{\cdot|Y} = \MaNormal{\tSyxiSxx}{V}{\inv{\tSxx}}$ comes from \eqref{eq:blr_posterior_case1}, and $\tSyx, \inv\tSxx$ are computed by evaluating their definitions with $\txi$. Notice that $Q$ can be equivalently expressed as the ELBO form\footnote{The symbol ``$\simeq$'' indicates equality up to an irrelevant additive constant.}:
\begin{align}
  Q(\xi;\txi) & \simeq \underbrace{\Exp[{A\sim\Pdf[\txi]{\cdot|Y}}]{\ln\Pdf[\xi]{Y|A}}}_{\eqdef Q_1(\xi;\txi)} - \underbrace{\KL{\Pdf[\txi]{A|Y}}{\Pdf[\xi]{A}}}_{\eqdef Q_2(\xi;\txi)} + \ln\pi(\xi)
\end{align}
where $Q_1-Q_2$ is just $\ELBO{Y}{\Pdf[\txi]{\cdot|Y}}{\opPdf_\xi}$ with $\Pdf[\txi]{\cdot|Y}$ acting as the variational posterior in ELBO. At the M-step the surrogate function is maximized (or increased) at $\xi$ such that $Q(\xi;\txi)\geq Q(\txi;\txi)$, and this produces a sequence $\set{\xi_n}_n$ that increases monotonically $\ln\Pdf{\xi|Y}$.

\paragraph*{E-Step}
We show in the next that the hyperparameters $\theta$ and DNN weights $\wt$ are separated into $Q_1$ and $Q_2$. The $Q_1$ term is
\begin{align}
  Q_1(\xi;\txi) & \simeq - \frac{\dmy}{2}\ln\abs{D} - \half\Exp[{A\sim\Pdf[\txi]{\cdot|Y}}]{\tracep{\inv V\paren{Y-A\Phi}\inv D\trpp{Y-A\Phi}}} \\
                & = - \frac{\dmy}{2}\ln\abs{D} -  \half\tracep{\inv V\paren{Y-\tSyxiSxx\Phi}\inv D\trpp{Y-\tSyxiSxx\Phi}}                       \\
                & \qquad -\half\tracep{\inv V \Exp[{A\sim\Pdf[\txi]{\cdot|Y}}]{\paren{A-\tSyxiSxx}\Phi \inv D \Phit\trpp{A-\tSyxiSxx}}}
\end{align}
and the expectation inside the last term evaluates to $\tracep{\Phi \inv D \Phit\inv\tSxx} V$, using the identity \eqref{eq:mvd_Normal_cov}. Therefore we get
\begin{align}
  Q_1(\xi;\txi) \simeq - \frac{\dmy}{2}\ln\abs{D} -  \half{\tracep{\inv V \tE \inv D \tE^\top}} - \frac{\dmy}{2} \tracep{\Phi \inv D \Phit\inv\tSxx}
\end{align}
with $\tE \defeq Y-\tSyxiSxx\Phi$, or equivalently in a sample-independent form:
\begin{align}
  \label{eq:Q1_summable_case1}
  Q_1(\wtp, \wts) \simeq -\half \sum_i \paren{\dmy\ln\sigma^2(x_i) + \sigma^{-2}(x_i)\paren{\tilde e_i^\top \inv V \tilde e_i + \dmy {\phi(x_i)}^\top\inv\tSxx\phi(x_i)}}
\end{align}
with $\tilde e_i \defeq y_i - \tSyxiSxx \phi(x_i)$. The $Q_2$ term can be computed using Lemma~\ref{lem:KL_normal}. Since $V$ is fixed through iterations, we obtain
\begin{align}
  \label{eq:Q2_MK_case1}
  Q_2(M,K) \simeq \frac{\dmy}{2} \ln\abs{K} + \frac{\dmy}{2}\tracep{\inv K\inv\tSxx} + \half\tracep{\inv V \tF \inv K \trp\tF}
\end{align}
with $\tF\defeq M-\tSyxiSxx$. This function is quadratic in $M$, and the minimizer for $K$ can be obtained by Corollary~\ref{coro:logdet_traceinv_minimizer} as
\begin{align}
  K = \frac{1}{\dmy}\paren{\dmy \inv\tSxx + \trp\tF\inv V \tF}.
\end{align}


\paragraph*{M-Step}
Notice that the objective $Q_1$ in \eqref{eq:Q1_summable_case1} is compatible with mini-batches training, so we apply first the SGD to optimize DNN weights $\wt$. On the other hand, $Q_2$ term admits closed-form solutions for $\theta$. The update scheme for $M,K$ is then determined on the basis of the regularization method proposed in Section~\ref{sec:regularize_ev_loss}. With the combined regularization, that is, by holding $M$ fixed and applying an Inverse Wishart hyperprior $\pi(K)=\InvWishart{K;\HypK}{\NdfK}$, the update is on $K$ only and reads
\begin{align}
  \label{eq:elbo_em_scheme_MK_fix_hyp}
  M = \tM, \ \ K = \frac{1}{\dmy+\NdfK}\paren{\dmy \inv\tSxx + \trp\tF\inv V \tF + \HypK}
\end{align}
Similarly, if only the hyperprior is applied, then $M$ is jointly updated and we have
\begin{align}
  \label{eq:elbo_em_scheme_MK_joint_hyp}
  M = \tSyxiSxx, \ \ K = \frac{1}{\dmy+\NdfK}\paren{\dmy \inv\tSxx + \HypK}
\end{align}
Variants to these updates can be easily obtained by taking into account the structural constraints on $K$. For instance, if $K$ is restricted to be diagonal, the $K$-update in \eqref{eq:elbo_em_scheme_MK_fix_hyp} and \eqref{eq:elbo_em_scheme_MK_joint_hyp} then needs to take the diagonal as well.

The EM algorithm is resumed in Algorithm~\ref{alg:ELBO_EM_framework}.

\begin{remark}
  The regularization is crucial for the update scheme to be well behaved. In fact, for the unregularized evidence $\ln\Pdf{Y|\xi}$, joint optimization of $M$ and $K$ leads to the update scheme of \eqref{eq:elbo_em_scheme_MK_flaw} which is pathological: the EM sequence converges to the degenerate MLE described in Theorem~\ref{thm:bll_mle_KVM}, as proved in Proposition~\ref{prop:conv_joint_scheme_unreg_ev}.
\end{remark}

\begin{algorithm}
  \caption{EM algorithm framework}\label{alg:ELBO_EM_framework}
  \begin{algorithmic}
    \Require $X, Y, V$, initialization $\xi_0=\set{\wt_0, M_0, K_0}$, maximum iteration $\Nmax$
    \Require fixed $M=M_0$ and/or hyperprior $\pi(K)=\InvWishart{K;\HypK}{\NdfK}$
    \State $n\leftarrow 0$
    \While{$n<\Nmax$, or not converged}
    \State   $\tilde\xi \leftarrow \xi_{n-1}$, that is, $\tilde\wt=\wt_{n-1}, \tM=M_{n-1}, \tK = K_{n-1}$
    \State   (E-step) Formulate $Q_1$ and $Q_2$ with $\txi$
    \State   (M-step for $\wt$) Maximization of \eqref{eq:Q1_summable_case1} wrt $\wt$ by SGD with mini-batches
    \State   (M-step for $\theta$) Update $M,K$ using
    \eqref{eq:elbo_em_scheme_MK_fix_hyp} or \eqref{eq:elbo_em_scheme_MK_joint_hyp}, or their variants with structural constraints
    \State $\xi_n\leftarrow$ the updated $\set{\wt, M, K}$
    \State $n\leftarrow n+1$
    \EndWhile
  \end{algorithmic}
\end{algorithm}

\subsection{Numerical Experiments}\label{sec:num_exp_case1}

\subsubsection{Bayesian Interpolation with heteroscedastic noise}\label{sec:em_bi_hetero}

\begin{figure}
  \centering
  \includegraphics[width=1.\textwidth]{{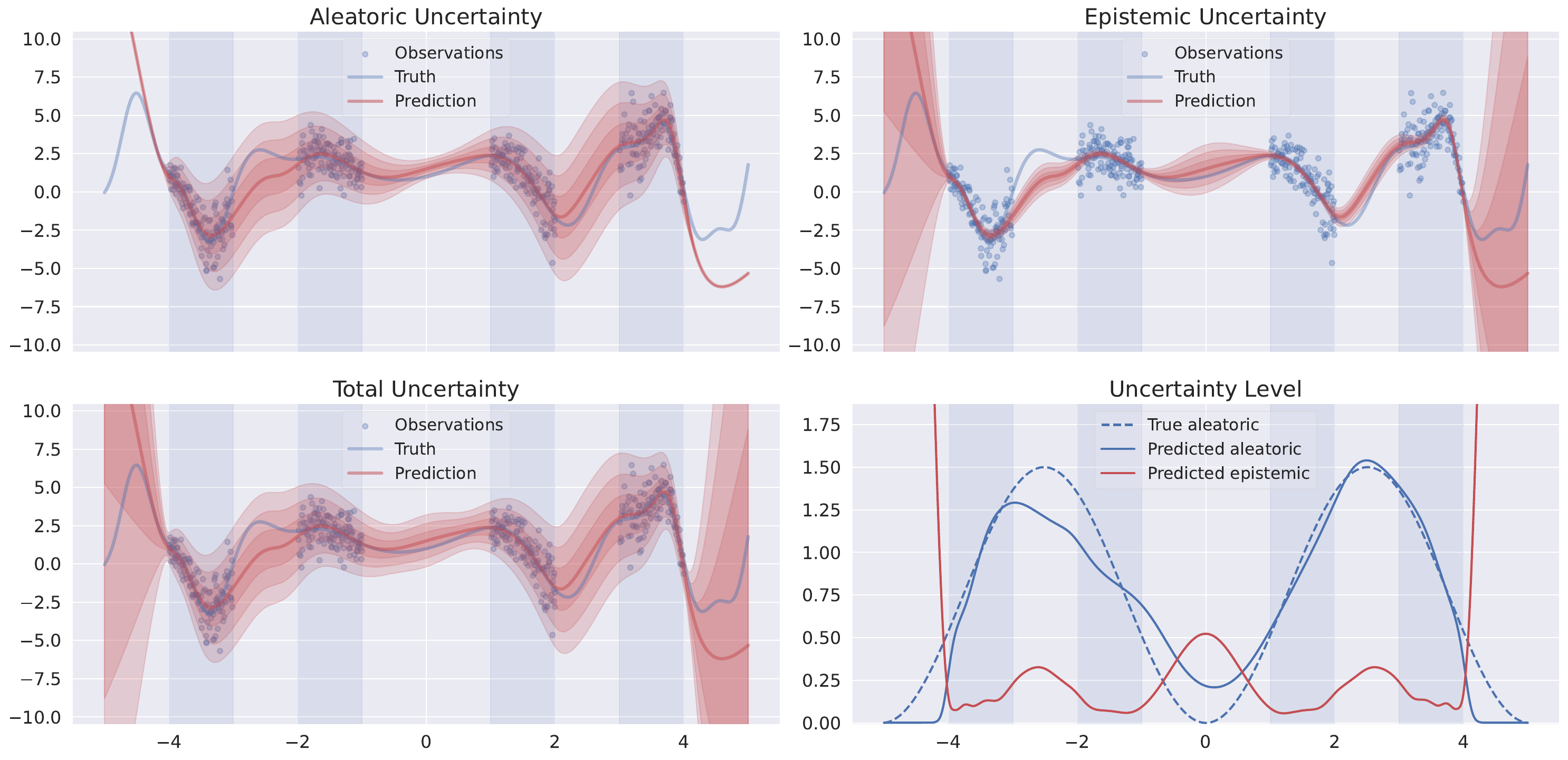}}
  \caption{EM algorithm~\ref{alg:ELBO_EM_framework} for Bayesian interpolation with DNN as basis function. We hold $M=0$ fixed and restrict $K$ to be isotropic. Color shades correspond to 1, 2, and 3 times of standard deviation.}
  \label{fig:em_dnn}
\end{figure}

First, we validate the EM algorithm through a 1d interpolation task. The scalar $V$ can be absorbed in the scaling $\sigma(\cdot)$, so is fixed to $1$. The ground truth here is a scalar function constructed from trigonometric functions, and the heteroscedastic scaling is modified from a squared cosine function. $N=500$ noisy observations are randomly and uniformly sampled on four disjoint intervals of length 1 around the location $[-3.5, -1.5, 1.5, 3.5]$. We randomly reserve 80\% of data for training and 20\% for validation. The whole setup is depicted in Figure~\ref{fig:em_dnn}.

Bayesian interpolation \citep{mackay_bayesian_1992} in classical literature is solved using fixed basis functions (\eg Gaussian kernels) under the setting of homoscedastic noise with diagonal $K$, which can be easily handled by our algorithm as special case as resumed in Algorithm~\ref{alg:ELBO_EM_Interpolation}. Here we parameterize $\phi(\cdot)$ and $\sigma(\cdot)$ using Multiple Layer Perceptron (MLP) with randomly initialized weights. The network of $\sigma$ shares the same backbone of $\phi$, which is composed of 4 layers with 64 units per layer and with \texttt{Softplus} as activation function, and the constant 1 is included in $\phi$ as bias. Algorithm~\ref{alg:ELBO_EM_framework} is applied by holding $M=0$ fixe. To improve convergence and reduce the complexity, $K$ is restricted to the isotropic covariance \ie $K=k I_\dmt$, and the scalar $k$ is updated by taking the average of diagonal in the $K$-update \eqref{eq:elbo_em_scheme_MK_fix_hyp}. The EM algorithm terminates when the relative change in $K$ and log-evidence both fall below $10^{-3}$.

\begin{figure}
  \centering
  \includegraphics[width=1.\textwidth]{{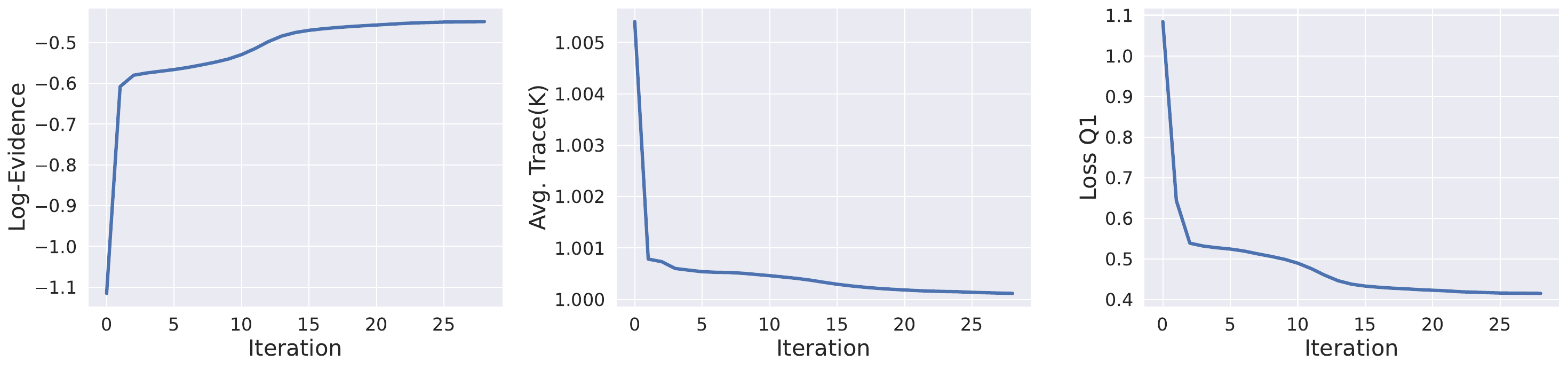}}
  \caption{Convergence of EM for the experiment in Figure~\ref{fig:em_dnn}. The algorithm terminates in less than 30 steps, with the relative changes below $10^{-3}$.}
  \label{fig:em_dnn_convergence}
\end{figure}

Figure~\ref{fig:em_dnn} shows the result of posterior predictive together with decomposed uncertainties. Pannels of the first row display the mean posterior predictive together with the aleatoric and epistemic uncertainty respectively. It can be observed that predicted aleatoric uncertainty (upper left), represented by the learned $\sigma$, closely captures heteroscedastic characteristics over observation intervals while attempting to track trends over the interpolation zones. On the other hand, the epistemic uncertainty (upper right) decays near data-dense zones. This aligns with Proposition~\ref{prop:asymptotic_epistemic}, where we proved epistemic uncertainty of BLL scales inversely with sample size. Moreover, the epistemic uncertainty explodes in regions far from observation intervals. The total uncertainty (lower left) is moderately elevated in observed and intrapolation regions (aleatoric-driven), but high in extrapolation regions (epistemic-driven), acting as a reliable protection against extrapolation. The profile of predicted uncertainties are plotted in the last pannel (lower right). The convergence of EM is depicted in Figure~\ref{fig:em_dnn_convergence}, in terms of the log-evidence, the value of $k$ and the loss $-Q_1$.

In this experiment, we have noticed that the end-to-end training of BLL models is computationally demanding and appears sensitive on architectural choices and other training parameters. As depicted in Figures~\ref{fig:em_dnn2} and \ref{fig:em_dnn32}, the resulting MLP models may have erratic behavior of uncertainties, manifesting as irregular and inconsistent profiles across interpolation zones, such as over-estimated aleatoric and under-estimated epistemic prediction. This can be explained by the learned bases, which are typically non-localized and ill-conditioned  with complex interference patterns. Figure~\ref{fig:learned_basis} shows a set of learned MLP bases $\phi$, which is highly redundant compared to a fixed bases of Gaussian kernels (condition number of order $10^8$ versus $10^2$). Adding weight decay regularization on DNN weights helps to some extent, but the learned bases still exhibit complex patterns that are hard to interpret. This suggests that further investigation on the role of DNN bases in BLL models will be necessary to improve training.



\subsubsection{Transfer learning}\label{sec:em_transfer_learning}
\paragraph*{Two-phase pipeline} Algorithm~\ref{alg:ELBO_EM_framework} provides a framework for uncertainty-aware transfer learning in deep neural networks. The methodology proceeds as follows. In the first phase, we pretrain \eg a probability density network on some source domain to learn simultaneously the representations and the noise scaling. Penultimate output layer of the pretrained representations then serve as $\phi$, and the final layer is replaced with a Bayesian linear layer. In the second phase, Algorithm~\ref{alg:ELBO_EM_framework} is then applied to the target domain for the optimization of BLL's parameters, with pretrained $\phi$ and $\sigma$ (either frozen or using a small learning rate). This approach avoids the full training of DNNs at each M-step of Algorithm~\ref{alg:ELBO_EM_framework}, and is computationally efficient compared to the fully end-to-end training.

\begin{figure}
  \centering
  \includegraphics[width=1.\textwidth]{{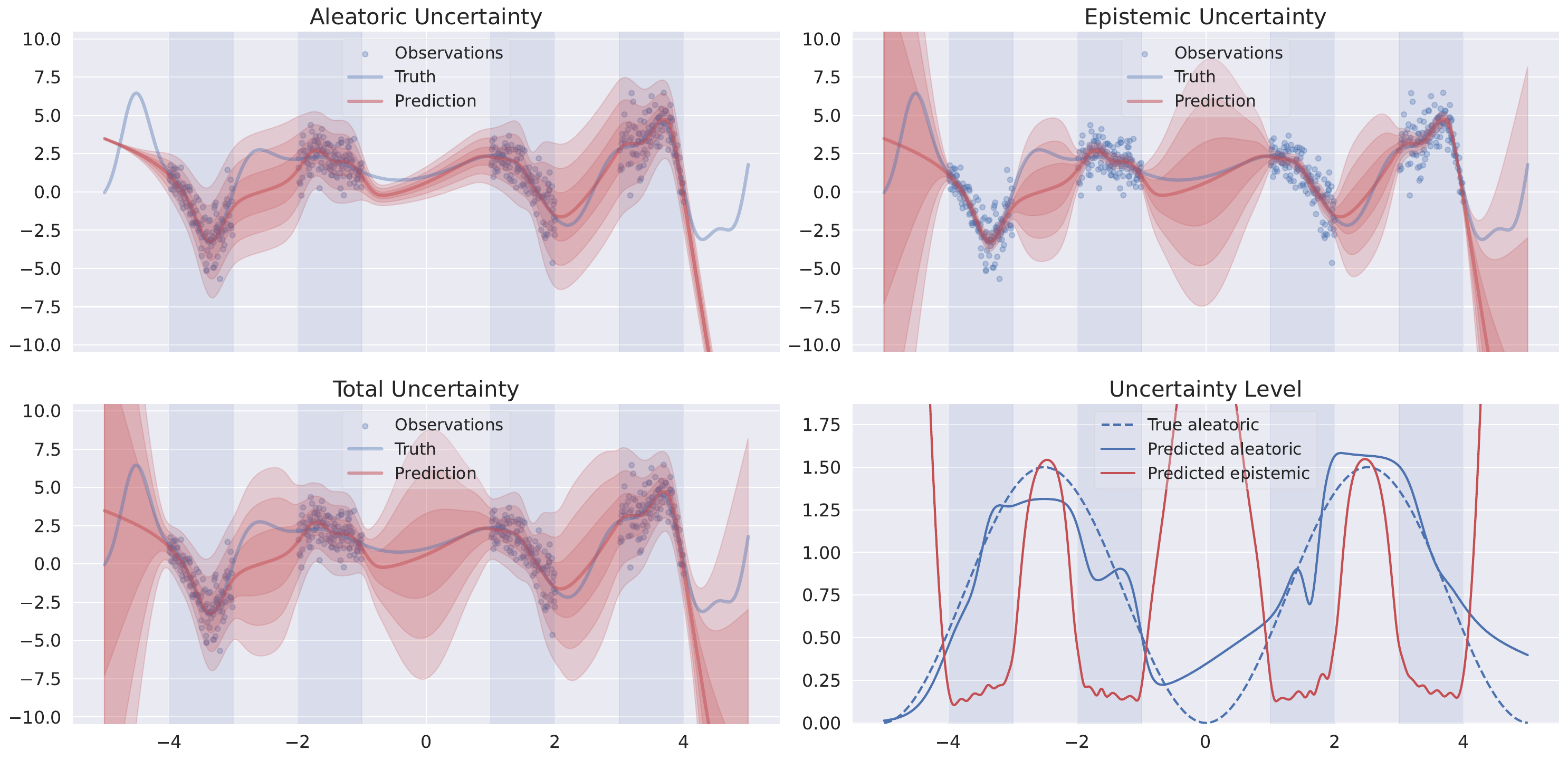}}
  \caption{Transfer learning approach for the experiment of Figure~\ref{fig:em_dnn}. Algorithm~\ref{alg:ELBO_EM_framework} is applied to adapt the BLL model with a fixed pretrained probability density network.
  }
  \label{fig:fixed_pretrained_density_network}
\end{figure}

Figure~\ref{fig:fixed_pretrained_density_network} demonstrates the two-phase pipeline with a fixed pretrained probability density network.  To improve the symmetry of predicted uncertainty in extrapolation zones, we have augmented the learned basis $\phi$ by adding the mirrored one $\tilde\phi:x\mapsto \phi(-x)$. Moreover, we used the Inverse Wishart hyperprior $\InvWishart{K;\HypK}{\NdfK}$ with the degree of freedom $\NdfK=1$
and the location parameter $\HypK=\invp{10^{-3} \cdot I+\Phi\inv D\Phi^\top}$, calculated from trained DNNs with a small regularization constant, which correspond to a scale invariant prior for $K$. The resulting epistemic uncertainty is considerably greater compared to Figure~\ref{fig:em_dnn}.

\begin{table}[H]
    \caption{Results for UCI regression tasks.
    }\label{tab:reg1results}

    \centering
    \resizebox{\textwidth}{!}{
        \begin{tabular}{c|cc|cc|cc}
            & \multicolumn{2}{c}{\textsc{Boston}} & \multicolumn{2}{c}{\textsc{Concrete}} & \multicolumn{2}{c}{\textsc{Energy}} \\
            & NLL ($\downarrow$) & RMSE ($\downarrow$)  & NLL ($\downarrow$) & RMSE ($\downarrow$)  & NLL ($\downarrow$)   & RMSE ($\downarrow$)  \\

            \hline

            MBLL scalar & ${2.78 \pm 0.37}$ & ${3.22 \pm 0.44}$  & ${3.18 \pm 0.11}$ & ${5.71 \pm 0.53}$ & ${2.39 \pm 0.22}$ & ${2.23 \pm 0.31}$ \\
            MBLL hyper & ${2.81 \pm 0.35}$ & ${3.19 \pm 0.41}$ & ${3.17 \pm 0.11}$ & ${5.66 \pm 0.50}$ & ${2.40 \pm 0.22}$ & ${2.20 \pm 0.30}$ \\
            VBLL     & ${2.55 \pm 0.06}$                   & ${2.92 \pm 0.12}$                  & ${3.22 \pm 0.07}$                   & ${5.09 \pm 0.13}$    & ${1.37 \pm 0.08}$    & ${0.87 \pm 0.04}$    \\
            LDGBLL   & $2.60\pm0.04$                       & $3.38\pm0.18$                         & ${2.97 \pm 0.03}$                & ${4.80 \pm 0.18}$ & $4.80 \pm 0.18$      & $0.50 \pm 0.02$      \\
            RBF GP   & ${2.41\pm0.06}$                  & ${2.83 \pm 0.16}$                  & $3.08 \pm 0.02$                     & $5.62 \pm 0.13$      & ${0.66 \pm 0.04}$ & ${0.47 \pm 0.01}$ \\
            Dropout  & ${2.36\pm0.04}$                  & ${2.78\pm0.16}$                    & ${2.90 \pm 0.02}$                & ${4.45 \pm 0.11}$ & $1.33 \pm 0.00$      & $0.53 \pm 0.01$      \\
            Ensemble & $2.48\pm0.09$                       & ${2.79\pm0.17}$                    & $3.04 \pm 0.08$                     & $4.55 \pm 0.12$      & ${0.58 \pm 0.07}$ & ${0.41 \pm 0.02}$ \\
            SWAG     & $2.64\pm0.16$                       & $3.08\pm0.35$                         & $3.19 \pm 0.05$                     & $5.50 \pm 0.16$      & $1.23 \pm 0.08$      & $0.93 \pm 0.09$      \\
            BBB      & ${2.39\pm0.04}$                  & ${2.74\pm0.16}$                    & $2.97 \pm 0.03$                     & $4.80 \pm 0.13$      & ${0.63 \pm 0.05}$ & ${0.43 \pm 0.01}$    \\
            \hline
            & \multicolumn{2}{c}{\textsc{Power}} & \multicolumn{2}{c}{\textsc{Wine}} & \multicolumn{2}{c}{\textsc{Yacht}}\\

            & NLL ($\downarrow$) & RMSE ($\downarrow$) & NLL ($\downarrow$) & RMSE ($\downarrow$) & NLL ($\downarrow$) & RMSE ($\downarrow$)\\

            \hline

            MBLL scalar & ${2.84 \pm 0.04}$ & ${4.12 \pm 0.16}$ & $1.04 \pm 0.04$ & $0.68 \pm 0.02$ & $1.73 \pm 0.57$ & $1.15 \pm 0.32$\\
            MBLL hyper & ${2.84 \pm 0.04}$ & ${4.13 \pm 0.16}$ & $1.04 \pm 0.04$ & $0.68 \pm 0.02$ & $1.76 \pm 0.61$ & $1.10 \pm 0.27$\\
            VBLL & ${2.73 \pm 0.01}$ & ${3.68 \pm 0.03}$ & $1.02 \pm 0.03$ & $0.65 \pm 0.01$ & $1.29 \pm 0.17$ & $0.86 \pm 0.17$\\
            LDGBLL & $2.77 \pm 0.01$ & $3.85 \pm 0.04$ & $1.02 \pm 0.01$ & $0.64 \pm 0.01$ & $1.13 \pm 0.06$ & $0.75 \pm 0.10$\\
            RBF GP & $2.76 \pm 0.01$ & $3.72 \pm 0.04$ & ${0.45 \pm 0.01}$ & ${0.56 \pm 0.05}$ & ${0.17 \pm 0.03}$ & ${0.40 \pm 0.03}$\\
            Dropout & $2.80 \pm 0.01$ & $3.90 \pm 0.04$ & $0.93 \pm 0.01$ & $0.61 \pm 0.01$ & $1.82 \pm 0.01$ & $1.21 \pm 0.13$\\
            Ensemble & ${2.70 \pm 0.01}$ & ${3.59 \pm 0.04}$ & $0.95 \pm 0.01$ & $0.63 \pm 0.01$ & $0.35 \pm 0.07$ & $0.83 \pm 0.08$\\
            SWAG & $2.77 \pm 0.02$ & $3.85 \pm 0.05$ & $0.96 \pm 0.03$ & $0.63 \pm 0.01$ & $1.11 \pm 0.05$ & $1.13 \pm 0.20$\\
            BBB & $2.77 \pm 0.01$ & $3.86 \pm 0.04$ & $0.95 \pm 0.01$ & $0.63 \pm 0.01$ & $1.43 \pm 0.17$ & $1.10 \pm 0.11$\\
            \hline
        \end{tabular}
    }
\end{table}
\begin{table}[H]
\caption{Results of NLEV and ECE.
}\label{tab:reg1results_nlev}
\centering
\resizebox{\textwidth}{!}{
\begin{tabular}{c|cc|cc|cc}
& \multicolumn{2}{c}{\textsc{Boston}} & \multicolumn{2}{c}{\textsc{Concrete}} & \multicolumn{2}{c}{\textsc{Energy}}\\

& NLEV ($\downarrow$) & ECE ($\downarrow$) & NLEV ($\downarrow$) & ECE ($\downarrow$) & NLEV ($\downarrow$) & ECE ($\downarrow$)\\

\hline

MBLL scalar & ${2.794 \pm 0.257}$ & ${0.049 \pm 0.015}$ & $3.270 \pm 0.099$ & $0.033 \pm 0.010$ & $5.075 \pm 0.336$ & $0.110 \pm 0.042$\\
MBLL hyper & ${2.736 \pm 0.285}$ & ${0.049 \pm 0.017}$ & $3.195 \pm 0.018$ & $0.029 \pm 0.012$ & $3.920 \pm 0.265$ & $0.097 \pm 0.042$\\
\hline
& \multicolumn{2}{c}{\textsc{Power}} & \multicolumn{2}{c}{\textsc{Wine}} & \multicolumn{2}{c}{\textsc{Yacht}}\\

& NLEV ($\downarrow$) & ECE ($\downarrow$) & NLEV ($\downarrow$) & ECE ($\downarrow$) & NLEV ($\downarrow$) & ECE ($\downarrow$)\\

\hline

MBLL scalar & ${2.850 \pm 0.039}$ & $0.014 \pm 0.004$ & ${1.057 \pm 0.036}$ & $0.014 \pm 0.005$ & $1.865 \pm 0.280$  & $0.113 \pm 0.054$\\
MBLL hyper & ${2.841 \pm 0.041}$ & ${0.014 \pm 0.007}$ & $1.042 \pm 0.037$ & $0.015 \pm 0.006$ & $1.630 \pm 0.366$ & $0.100 \pm 0.043$\\
\hline

\end{tabular}
}

\end{table}

\subsubsection{UCI datasets} \label{sec:uci_datasets}

We demonstrate the performance of our BLL models on UCI regression datasets, previously investigated in the studies \citep{watson_latent_2021,harrison_variational_2024}. The precedent experimental procedure is followed here as closely as possible.

We test the two-phase transfer learning pipeline with the homoscedastic model (\ie $\sigma\propto 1$), and a DNN feature mapping $\phi$ composed of 2 hidden layers of 50 units, with \texttt{leaky relu} activation.
A probability density network is first pretrainned (the constant scaling $\sigma$ is jointly learned with $\phi$) with the AdamW optimizer (with a weight decay of 0.01 and a learning rate adapted to the validation loss, and the gradient clip of value 1.0). Then BLL hyperparameters together with the constant $\sigma$ are estimated with the EM algorithm with frozen weights of $\phi$ and pretrained $\sigma$ as initialization. The M-step update for constant $\sigma$ is given in Algorithm~\ref{alg:ELBO_EM_Interpolation}.


More specifically, we test the two regularization techniques of Section~\ref{sec:regularize_ev_loss}. In the first tehcnique we hold $M=0$ fixed and contraint $K$ to be isotropic (termed as ``MBLL scalar'' in the results). In the second technique we place an hyperprior $\InvWishart{K;I}{1}$ and contraint $K$ to be diagonal, and use the pretrained last layer weights of the mean network as initialization for $M$ (termed as ``MBLL hyper'' in the results). The EM algorithm stops when the relative change in $M, \sigma, K$ and log-evidence all fall below $10^{-4}$.

We randomly initialize DNN weights and split data as 72\%, 18\%, 10\% for training, validation, and test respectively. We run 20 seeds and report the values of RMSE and negative log-likelihood (NLL) on the test set. Table~\ref{tab:reg1results} summarizes the results of our BLL model with the two regularization techniques compared to several baselines\footnote{The results of baselines are taken from \citep{harrison_variational_2024} which are replicated from \citep{watson_latent_2021}.} including VBLL \citep{harrison_variational_2024}, Latent Derivative BLL (LDGBLL) \citep{watson_latent_2021}, Gaussian process with RBF kernel (RBF GP), Monte Carlo Dropout \citep{gal_dropout_2016}, Deep Ensemble \citep{lakshminarayanan_simple_2017}, SWAG \citep{maddox_simple_2019}, Bayes by Backprop (BBB) \citep{blundell_weight_2015}.

The negative log-evidence (NLEV, computed from \eqref{eq:blr_evidence_case1}) and expected calibration error (ECE) are reported for BLL models in Table~\ref{tab:reg1results_nlev}. To our knowledge, these metrics have not been reported for the baseline models in previous works. An instance of uncertainties and calibration curve on the test data of dataset \textsc{Boston} is shown in Figure~\ref{fig:uq_cali_boston}.

On these datasets, our BLL models have worse performance
\footnote{The \textsc{Energy} dataset is the only one with multivariate output ($p=2$). We calculated the metrics based on the vector norm, which could explain the significant difference observed compared to other baseline methods.}
compared to most baselines. This may be explained by the fact that the learned DNN features from the pretrained network, which is frozen during the EM stage, may not be optimal for the BLL model.Nonetheless, the ECE values on most datasets are below $10\%$, indicating that the predicted uncertainties are well calibrated. While the transfer learning approach is computationally efficient, further investigation is needed to improve its performance on BLL models.



\section{Extension to Matrix-T Distribution}
\label{sec:mvblr_T}
The BLL framework developed in Section~\ref{sec:mvblr} assumes known baseline noise covariance $V$. In practice, one could estimate $V$ from data, which introduces a source of uncertainty that must be taken into account by a Bayesian approach.
To address this limitation, we now generalize the previous framework to handle the more realistic case where $V$ is unknown and treated as a random variable.

\subsection{Bayesian Regression}\label{sec:bayesian_regression_T}
Specifically, we assume an Inverse Wishart prior for the unknown matrix $V$:
\begin{align}
  \label{eq:blr_prior_V_case2}
  V\sim\InvWishart{\Spr}{\Ndf}
\end{align}
where $\Ndf > 2\dmy$ is the degree of freedom and $\Spr$ is a fixed symmetric positive definite matrix. The corresponding density function is given by
\begin{align}
  \InvWishart{V;\Spr}{\Ndf} \propto
  {\abs{\Spr}^{\frac{\Ndf-\dmy-1}{2}}}{\abs{V}^{-\frac \Ndf 2}}\exptr\paren{-\half V^{-1}\Spr}
\end{align}
It has the mean $\Spr/\paren{\Ndf-2\dmy-2}$ as long as $\Ndf>2\dmy+2$ and the mode $\Spr/\Ndf$. Conditioned on $V$, the prior for $A$ and the likelihood remain the same as in the known $V$ case. The joint distribution of $[V\ A\ Y]$ is expressed then as the product of the likelihood $Y|(A,V)$, the conditional prior $A|V$, and the prior for $V$:
\begin{align}
  \label{eq:blr_joint_case2}
  [V\ A\ Y] \sim \MaNormal{A\Phi}{V}{D} \otimes \MaNormal{M}{V}{K} \otimes \InvWishart{\Spr}{\Ndf}
\end{align}

The matrix-T distribution is the matrix-variate extension of the classical multivariate-T distribution and arises naturally due to the conjugacy of the Normal-Inverse Wishart prior.
A matrix $T\sim p\times n$ that follows a matrix-T distribution with mean $M$, row and column covariances $\Sigma$ and $\Omega$, and degrees of freedom $\nu$, has the density
\begin{align}
  \MaStudent{T;M}{\Sigma}{\Omega}{\nu} \propto \abs{\Sigma}^{-\frac n 2}\abs{\Omega}^{-\frac p 2} \abs{I_p+\Sigma^{-1}\paren{T-M}\Omega^{-1}\paren{T-M}^\top}^{-\paren{\frac{\nu+n+p-1}{2}}}
\end{align}

In parallel to \eqref{eq:mvd_Normal_cov}, it holds for any square matrix $Q$ of appropriate dimensions and for $\nu>2$ that
\begin{align}
  \label{eq:mvd_T_cov}
  \Exp{\paren{T-M} Q \paren{T-M}^\top} = \frac{1}{\nu-2}\trace\paren{Q^\top \Omega} \Sigma.
\end{align}
For the proof see Theorem~4.3.2 of \citep{gupta_matrix_1999}.

\paragraph*{Helper identities}
In addition to the notation used in Section~\ref{sec:mvblr}, we introduce new symbols $\Syx$, $\Sycx$, and $\Sycxa$, as summarized in Table~\ref{tab:symbols}. These new terms satisfy the following helper identities:
\begin{align*}
  \Sycx &= \paren{Y-M\Phi}\iPxx \paren{Y-M\Phi}^\top \\
  \Sycxa &= \paren{Y-A\Phi}\inv D\trpp{Y-A\Phi} + \paren{A-M}\inv K\trpp{A-M} \\
  &= [(A-M) \ (Y-M\Phi)] {\PsiMatKX}^{-1}
  \begin{bmatrix}
    \paren{A-M}^\top\\
    \paren{Y-M\Phi}^\top
  \end{bmatrix}
\end{align*}

Theorem~\ref{thm:blr_case2} below is the counterpart of Theorem~\ref{thm:blr_case1} for the case $V$ is unknown. It establishes among others that the joint $[A\ Y]$, the marginals as well as the posterior predictive all follow matrix-T distributions.

\begin{theorem}[Bayesian regression: $V$ unknown]
  \label{thm:blr_case2}
  Suppose that the noise covariance $V$ follows \eqref{eq:blr_prior_V_case2}. With the symbols defined in Table~\ref{tab:symbols}, it holds the following distributions:
  \begin{enumerate}
    \item  The joint distribution of $[A\ Y]$ is
      \begin{align}
        \label{eq:blr_joint_AY_case2}
        [A\ Y] \sim \MaStudent{[M\ M\Phi]}{\Spr}{\PsiMatKX}{\Ndf-2\dmy}
      \end{align}
    \item
      The evidence is
      \begin{align}
        \label{eq:blr_evidence_case2}
        Y \sim \MaStudent{M\Phi}{\Spr}{\Pxx}{\Ndf-2\dmy}
      \end{align}
      and the marginal of $A$ is
      \begin{align}
        \label{eq:blr_marginal_A_case2}
        A \sim \MaStudent{M}{\Spr}{K}{\Ndf-2\dmy}
      \end{align}
    \item
      The posterior $[A \ V]|Y$ follows a compound Normal-Inverse Wishart distribution with the marginals
      \begin{align}
        &
        \begin{aligned}
          \label{eq:blr_post_marginal_A_case2}
          A|Y \sim \MaStudent{\Syx\iSxx}{\Spr+\Sycx}{\iSxx}{\Ndf+\dmN-2\dmy}
        \end{aligned}\\
        &
        \begin{aligned}
          \label{eq:blr_post_marginal_V_case2}
          V|Y \sim \InvWishart{\Spr+\Sycx}{\Ndf+\dmN}
        \end{aligned}
      \end{align}
      Moreover, conditioned additionally on $V$, the posterior of $A$ becomes
      \begin{align}
        \label{eq:blr_post_update_A_case2}
        A|Y,V \sim \MaNormal{\Syx\iSxx}{V}{\iSxx}
      \end{align}
      which is the same as \eqref{eq:blr_posterior_case1}, and conditioned additionally on $A$, the posterior of $V$ becomes
      \begin{align}
        \label{eq:blr_post_update_V_case2}
        V|Y,A \sim \InvWishart{\Spr+\Sycxa}{\Ndf+\dmN+\dmx}
      \end{align}
    \item
      Given $L$ new data points $\Phi_*\sim \dmx\times L$, the posterior predictive is
      \begin{align}
        \label{eq:blr_predictive_case2}
        Y_*|Y \sim \MaStudent{\Syx\iSxx \Phi_*}{\Spr+\Sycx}{{D_*}+\Phi_*^\top\iSxx \Phi_*}{\Ndf+\dmN-2\dmy}
      \end{align}
      Moreover, the posterior predictive knowing additionally $A$ is
      \begin{align}
        \label{eq:blr_predictive_case2_A}
        Y_*|Y, A \sim \MaStudent{A \Phi_*}{\Spr+\Sycxa}{D_*}{\Ndf+\dmN+\dmx-2\dmy}
      \end{align}
      which is different from $Y_*|A$ since
      \begin{align}
        \label{eq:blr_predictive_case2_A_only}
        Y_*|A \sim \MaStudent{A \Phi_*}{\Spr+\paren{A-M}K^{-1}\trpp{A-M}}{D_*}{\Ndf+\dmx-2\dmy}.
      \end{align}
  \end{enumerate}
\end{theorem}

\subsubsection{Quantification of uncertainties}
In parallel to the uncertainty decomposition for the case where $V$ is known, now we present the following quantification result.

\begin{proposition}[Decomposition of uncertainties: $V$ unknown]
  \label{prop:blr_UQ_case2}
  Let $\lambda=\set{A,V}$ and $\theta=\set{M,K,\Ndf,\Spr}$ be the likelihood and prior parameters as in Section~\ref*{sec:blr_UQ_case1}. Given the data $Y$, the aleatoric and epistemic uncertainty of the predictive $Y_*$ are respectively
  \begin{subequations}
    \label{eq:blr_uncertainties_case2}
    \begin{align}
      \Exp{\Var{Y_*|\lambda}|Y,\theta} &= \frac{\trace\paren{D_*}}{\Ndf+\dmN-2\dmy-2} \paren{\Spr+\Sycx} \label{eq:blr_uncertainties_case2_aleatoric} \\
      \Var{\Exp{Y_*|\lambda}|Y,\theta} 
      &= \frac{\trace\paren{\Phi_*^\top \iSxx \Phi_*}}{\Ndf+\dmN-2\dmy-2}\paren{\Spr+\Sycx},\label{eq:blr_uncertainties_case2_epistemic}
    \end{align}
  \end{subequations}
\end{proposition}

\subsection{Parameter Estimation}
Following the same evidential framework as in Section~\ref{sec:blr_parm_estim}, we estimate the hyperparameters $\set{M, K, \Spr}$ by maximizing the evidence function. The degree of freedom $\Npr$, on the other hand, is fixed to some small number \eg $\Npr=2\dmy+1$. We introduce the following shorthands, in addition to those in \eqref{eq:ev_loss_case1}:
\begin{align}
  \label{eq:shorthand_H_Cv}
  H \defeq \Spr + E \iPxx E^\top, \ \Cn\defeq \frac{\Gamma_p\paren{\frac{\Nprp+\dmN}{2}}}{\Gamma_p\paren{\frac{\Nprp}{2}}\pi^{\frac{\dmN\dmy}{2}} }, \ \Nprp\defeq \Npr-\dmy-1
\end{align}
where $\Gamma_p$ denotes the multivariate gamma function. The loss function to be minimized is the negative log-evidence~\eqref{eq:blr_evidence_case2} and can be expressed equivalently as, up to some irrelevant constant:
\begin{align}
  \label{eq:ev_loss_case2}
  \lL(M, K, \Spr) \defeq - \ln \Cn - \frac{\Nprp}{2} \ln\abs{\Spr} + \frac{\dmy}{2}\ln\abs{\Pxx} + \frac{\Nprp+\dmN}{2} \ln\abs{H}
\end{align}
Next, in the same vein as Theorem~\ref{thm:bll_mle_KVM}, we establish the unique global minimizer for the joint optimization of hyperparameters by using the criteria of the first order optimality. Notably it shows again that \eqref{eq:bll_mle_MK_case1} is a degenerate global minimizer.

\begin{theorem}[Joint minimization: $V$ unknown]
  \label{thm:bll_mle_KVM_T}
  Under the same assumptions of Theorem~\ref{thm:bll_mle_KVM}, the following statements hold:
  \begin{enumerate}
    \item
      The joint minimization of the loss with respect to $(M,K,\Spr)\in\lM_{\dmy,\dmt} \times \Spp^\dmt \times \Spp^\dmy$ is reduced to the minimization of $K\in\Spp^\dmt$ alone:
      \begin{align}
        \min_{M,K,\Spr} \lL(M,\Spr,K) = \min_K \lL(\hat M, \hat\Spr, K)
      \end{align}
      where the unique minimizer $\hat M$ is the same as in \eqref{eq:bll_mle_MK_case1}, and $\hat\Spr$ is $K$-independent
      \begin{align}
        \label{eq:bll_mle_V_case2}
        \hat \Spr &= \frac{\Nprp}{\dmN} Y \hinv D P \hinv D Y^\top.
      \end{align}
      Here $P$ is the same orthogonal projector defined in \eqref{eq:projector_P_PhiD}.
    \item
      The estimator $\hat \Spr$ is biased
      \begin{align}
        \Exp{\what \Spr} = \paren{\frac{\dmN-\dmt}{\dmN} \cdot \frac{\Nprp}{\Nprp-\dmy-1}} \Spr.
      \end{align}
    \item
      With $\hat M$ and $\hat\Spr$, the loss \eqref{eq:ev_loss_case2}
      is concave and strictly increasing in $K$. In particular the singular point $\hat K=0$ is the unique global minimizer.
  \end{enumerate}
\end{theorem}

The regularization techniques of Section \ref{sec:regularize_ev_loss} can be applied similarly in this case to stabilize the evidential loss.

\subsection{Optimization with EM}\label{sec:elbo-em-T}
We apply the same EM framework of Section~\ref{sec:EM_algo_case1} for estimation of hyperparameters, with the unobserved quantities $\lambda=\set{A, V}$. Let $\xi=\theta \cup \wt = \set{M, \Spr, K, \wtp, \wts}$, and $\txi = \tilde\theta \cup \tilde\wt = \set{\tM, \tSpr, \tK, \twtp, \twts}$. For simplicity, we hold $M$ fixed and omit the hyperprior on $K$ in the following derivations. Following the same notations of Section~\ref{sec:EM_algo_case1}, the surrogate function is then
\begin{align}
  Q(\xi;\txi) & = \Exp[{[A,V]\sim\Prb[\txi]{\cdot|Y}}]{\ln\Prb[\xi]{[A, V, Y]}} \notag \\
  & \simeq {\Exp[{V\sim\Prb[\txi]{\cdot|Y}}]{\Exp[{A\sim\Prb[\txi]{\cdot|V,Y}}]{\ln\Prb[\xi]{Y|A,V}}}} - {\KL{\Prb[\txi]{[A,\ V]|Y}}{\Prb[\xi]{[A,\ V]}}} \\
  & = Q_1(\xi;\txi) - Q_2(\xi;\txi)
\end{align}
where the posterior
\begin{align}
  \Prb[\txi]{[A,V]|Y} = \InvWishart{V;\tSpr+\tSycx}{\Npr+\dmN} \times \MaNormal{A;\tSyxiSxx}{V}{\inv{\tSxx}}
\end{align}
comes from \eqref{eq:blr_post_marginal_V_case2}.

For a $\dmy$-dimensional covariance matrix $W\sim\InvWishart{\Spr}{\Npr}$, it holds $\Exp{\inv W} = \paren{\Npr-p-1}\inv\Spr$. Using this fact we obtain the $Q_1$ term from the matrix-Normal case in Section~\ref{sec:EM_algo_case1}.
It holds that
\begin{align}
  Q_1 \simeq - \frac{\dmy}{2} \ln\abs{D} - \frac{\Nprp+N}{2}\tracep{\invp{\tSpr+\tSycx} \tE\inv D\tE^\top} - \frac{\dmy}{2} \tracep{\Phi \inv D\Phi^\top \inv\tSxx}
\end{align}
or equivalently in a separable form:
\begin{align}
  \label{eq:Q1_summable_case2}
  Q_1 \simeq -\half \sum_i \paren{\dmy\ln\sigma^2(x_i) + \sigma^{-2}(x_i)\paren{\paren{\Nprp+\dmN} \tilde e_i^\top \invp{\tSpr+\tSycx} \tilde e_i + \dmy {\phi(x_i)}^\top\inv\tSxx\phi(x_i)}}.
\end{align}

The following elementary result will be used for the simplification of the $Q_2$ term:
\begin{lemma}
  \label{lem:KL_student}
  Let $V\sim\dmy\times\dmy$ follow an Inverse Wishart distribution $\InvWishart{\Spr}{\Npr}$ and $A\sim\dmy\times\dmx$ conditioned on $V$ follow a matrix Normal distribution $\MaNormal{M}{V}{K}$.
  Denote by $\xi=\set{M,K,\Spr,\Npr}$ and $\txi=\set{\tM,\tK,\tSpr, \tNpr}$. Then the KL-divergence between the joint distribution $\Prb[\txi]{A, V}$ and $\Prb[\xi]{A, V}$ equals to, up to some constant which is independent of $\xi$:
  \begin{align}
    \KL{\Prb[\txi]{[A, V]}}{\Prb[\xi]{[A, V]}} & \simeq \frac\dmy 2 \paren{\ln\abs{K} + \tracep{\inv K \tK}} + \\
    &\frac{\tNprp}{2}\trace\set{\inv \tSpr\paren{\paren{M-\tM}\inv K\trpp{M-\tM} + \Spr}} - \frac{\Nprp}{2}\ln\abs{\Spr}
  \end{align}
  with $\tNprp \defeq \tNpr-p-1, \Nprp\defeq\Npr-p-1$.
\end{lemma}
Applying this on $Q_2$ with the prior $\Prb[\xi]{[A, V]}=\InvWishart{V;\Spr}{\Npr}\times\MaNormal{A;M}{V}{K}$ gives
\begin{align}
  Q_2(M,K,\Spr)
  \simeq \frac{\dmy}{2} \ln\abs{K} & + \frac{\dmy}{2} \tracep{\inv K \inv{\tSxx}} - \frac{\Nprp}{2}\ln\abs{\Spr} + \\
  & \frac{\Nprp+\dmN}{2}\trace\paren{\invp{\tSpr+\tSycx} \paren{\tF \inv K \trp \tF + \Spr}}
\end{align}
By Corollary~\ref{coro:logdet_traceinv_minimizer}, $Q_2$ as function of $K$ admits the unique minimizer
\begin{align}
  \label{eq:elbo_em_scheme_K_case2}
  K = \inv\tSxx+\paren{\frac{\Nprp+\dmN}{\dmy}}\trp\tF\invp{\tSpr+\tSycx}\tF
\end{align}
On the other hand, $Q_2$ is convex in $\Spr$, with the unique minimizer
\begin{align}
  \label{eq:elbo_em_scheme_V_case2}
  \Spr = \frac{\Nprp}{\Nprp+\dmN}\paren{\tSpr+\tSycx}
\end{align}
Structural constraint can be put on $\Spr$ and $K$, in the same vein as for Algorithm~\ref{alg:ELBO_EM_Interpolation}. In that case the update \eqref{eq:elbo_em_scheme_K_case2} and \eqref{eq:elbo_em_scheme_V_case2} have to be adapted.

Finally, $Q_2$ is quadratic in $M$ and the unique minimizer $M = \tSyxiSxx$ would result in the same pathological joint scheme \eqref{eq:elbo_em_scheme_MK_flaw}. EM algorithm with fixed $M$ is summarized in Algorithm~\ref{alg:ELBO_EM_framework_M_fixed_case2}. The inclusion of hyperprior and joint update of $M$ is straightforward, as done in Section~\ref{sec:EM_algo_case1}.

\begin{algorithm}
  \caption{EM algorithm framework ($M$ fixed): $V$ unknown}\label{alg:ELBO_EM_framework_M_fixed_case2}
  \begin{algorithmic}
    \Require $X, Y, M, \Npr$, initialization $\xi_0=\set{\wt_0, K_0, \Spr_0}$, maximum iteration $\Nmax$
    \State $n\leftarrow 0$
    \While{$n<\Nmax$}
    \State   $\tilde\xi \leftarrow \xi_{n-1}$, that is, $\tilde\wt=\wt_{n-1}, \tK = K_{n-1}, \tSpr=\Spr_{n-1}$
    \State   (E-step) Formulate $Q_1$ with $\txi$
    \State   (M-step for $\wt$) Maximization of \eqref{eq:Q1_summable_case2} wrt $\wt$ by SGD with mini-batches
    \State   (M-step for $\wt$) Optimization of $\wt$ by applying \eg SGD with mini-batches on $Q_1$.
    \State   (M-step for $\theta$) Update $K$ using \eqref{eq:elbo_em_scheme_K_case2}, and $\Spr$ using \eqref{eq:elbo_em_scheme_V_case2} or variants with structural constraints
    \State $\xi_n\leftarrow$ the updated $\set{\wt, K, \Spr}$
    \State $n\leftarrow n+1$
    \EndWhile
  \end{algorithmic}
\end{algorithm}

\subsection{Numerical Experiments}\label{sec:num_exp_case2}
We demonstrate our framework on multivariate time series forecasting using Beijing air quality data from \citep{misc_beijing_multi-site_air-quality_data_501}. The dataset, after standardization (centering and dividing by standard deviation), comprises 4 years of hourly measurements (March 2013 onward). The model incorporates Inputs: 5 inputs of meteorological condition (temperature, pressure, dew point, precipitation, wind speed) and 6 outputs of pollutant concentration (PM2.5, PM10, SO2, NO2, CO, O3).

We formulate a nonlinear $(P,Q)$-VARX (Vector Autoregressive with Exogenous Variable) model:
\begin{align}
y_t = A \phi(\tilde x_t) + \sigma(\tilde x_t)\varepsilon_t, \quad \tilde x_t \defeq [x_t,\ldots,x_{t-P+1}, y_{t-1},\ldots,y_{t-Q}]
\end{align}
where $P$ and $Q$ define the input and autoregressive orders. We take $P=1, Q=2$ \footnote{Other VARX orders \eg $P=1, Q=1$ or $P=2, Q=2$ have been tested and give similar results. The best order can be selected within the evidential framework proposed in Section~\ref{sec:ev_framework}.}
 which yields the corresponding regression model's input dimension $\dmx=17$.

\paragraph*{Two-phase pipeline} BLL priors are imposed on the coefficient matrix $A$ and noise covariance $V$. We applied the two-phase pipeline from Section~\ref{sec:em_transfer_learning}:
\begin{enumerate}
  \item Feature Learning. Train a probability density network $\mu(\cdot), \sigma(\cdot)$ (same MLP architecture from Section~\ref{sec:em_bi_hetero} but with 128 units per layer) by minimizing the loss $\sum_i \ell(y_i,\tilde x_i)$ with
  \begin{align}
    \ell(y,\tilde x) \defeq \sigma^{-2}(\tilde x) \trpp{y-\mu(\tilde x)}\inv {V_0} \paren{y-\mu(\tilde x)} + \ln\sigma^2(\tilde x)
  \end{align}
  where the empirical covariance matrix $V_0$ is estimated from the training output data. We use first 3 years data for training, next 6 months for validation.
  \item BLL Adaptation. Extract and freeze the feature map $\phi(\cdot)$ (penultimate layer of $\mu$) and noise scale $\sigma(\cdot)$, then estimate BLL parameters via Algorithm~\ref{alg:ELBO_EM_framework_M_fixed_case2} on $\dmN=1000$ random training points (to reduce the computation load).
\end{enumerate}

\begin{figure}[H]
  \centering
  \includegraphics[width=1.\linewidth]{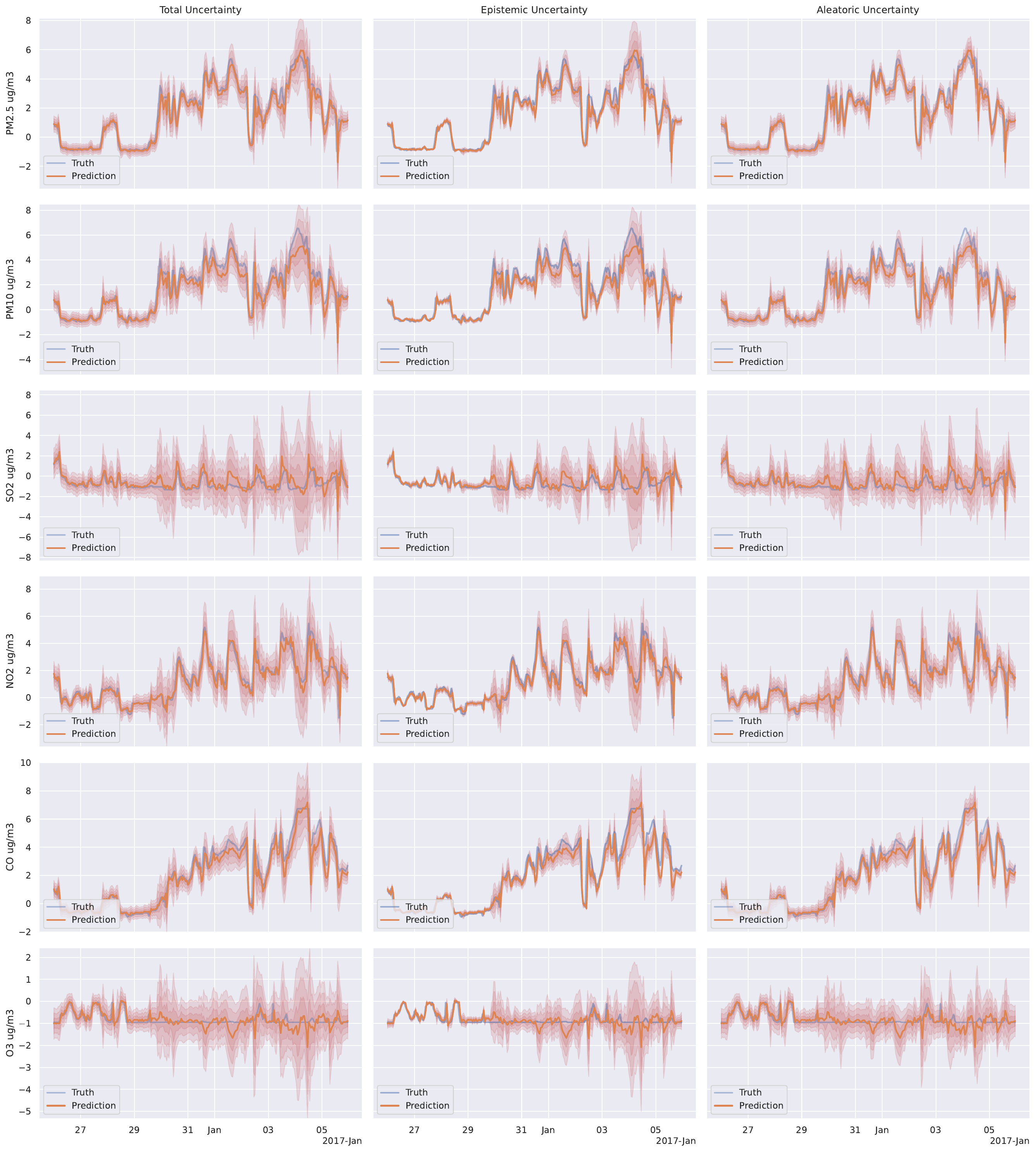}
  \caption{Predictions of Beijing air quality using a nonlinear (1,2)-VARX BLL model, with $\Npr=15$ and an isotropic $K$ and $\Spr$. Negative values are due to data standardization.}
  \label{fig:beijing}
\end{figure}

We fix $M$ to the constant unit vector, and set the degree of freedom to $\Npr=2\dmy+3=15$ (the result seems robust to this parameter). Isotropic and diagonal $K$ and $\Spr$ have been tested both showing similar results.
The prediction as well as the uncertainty profiles for the period of test on Dec. 26, 2016 to Jan. 05, 2017 are displayed in Figure~\ref{fig:beijing}. The model reveals a distinct behavioral shift in pollutant concentrations during the New Year period (December 30 to January 5). This anomaly likely stems from
specific human activities during holidays (\eg fireworks and reduced industrial emissions).  As these conditions occur infrequently compared to typical days, they represent a distribution shift in the data space. The results successfully capture this phenomenon through its uncertainty decomposition:
\begin{itemize}
  \item Epistemic uncertainty rises significantly above baseline levels, correctly identifying the period as out-of-distribution;
  \item Aleatoric uncertainty shows only modest increase, reflecting that measurement noise characteristics remain relatively stable.
\end{itemize}
More results on the period Jan. 10 to Jan. 20, 2017 is displayed in Figure~\ref{fig:beijing_3}.


\section{Discussion and Conclusion}\label{sec:conclusion}

This work has established the Bayesian Last Layer framework for uncertainty-aware multivariate regression in heteroscedastic settings. By providing closed-form expressions for both aleatoric and epistemic uncertainties, our method enables efficient uncertainty quantification through a single forward pass, overcoming the computational bottlenecks of sampling-based approaches.

We provided theoretical analysis of the evidential framework for parameter estimations, and revealed a critical pathology of the evidential loss: joint optimization of the BLL hyperparameters leads to a degenerate MLE solution. With regularization, the specialized EM algorithm offers unbiased gradient estimation on mini-batches and straightforward adaptation for transfer learning scenarios. The whole BLL framework was extended to the matrix T-distribution where the noise covariance is unknown. Numerical experiments demonstrates BLL's ability to deliver uncertainty estimates while maintaining the predictive accuracy.

The study also reveals important limitations in the current model that require further investigation.
\begin{itemize}
  \item We assumed a restrictive scalar scaling $\sigma^2(x)V$ for noise covariance --- a tradeoff made to mitigate the matrix-variate modeling and the heteroscedasticity --- that limits the expressiveness of BLL. How to introduce in BLL framework the true matrix-variate heteroscedasticity of type $V(x)$ remains a challenge. The recent approach in \citep{harrison_heteroscedastic_2025} could be a promising direction.
  \item Feature learning is decoupled from Bayesian inference in the analysis. However, the observed sensitivity of predicted uncertainty to feature representation in the numerical results strongly suggests that feature learning should be regularized in some way, in order to better balance the stability of fixed basis and the adaptability of learned representations.
  \item The choice of hyperprior for $K$ has a significant impact on the model's performance. More systematic investigation into alternative prior structures, such as hierarchical or sparsity-inducing priors, could enhance the flexibility and robustness of the BLL framework.
  \item Numerical instabilities arising from ill-conditioned matrices occasionally disrupt the training procedure (this is also related to the learned representation which is highly redundant). While the hyperprior, diagonal and scalar constraints for $K$ and $\Spr$ used in this work provide partial mitigation, more sophisticated regularization strategies could offer improved stability.
  \item As we pointed out, the Normal distribution based BLL model can be viewed as a  Gaussian Process with linear kernel $k(x,x') = \phi(x)^\top K \phi(x')$, which naturally connects it to the literature of Deep Kernel Learning \citep{wilson_deep_2016}. By analogy, when Student-T distributions are used, the resulting BLL model could be interpreted as a Student-T process \citep{shah_student-t_2014,xu_sparse_2024}. This provides new insights on extending the BLL framework to enhance its flexibility and modeling capacity.
\end{itemize}

\paragraph*{Code and Data Availability}
For reproducible research, we provide the Python package \texttt{rundle} with notebooks and some datasets, available at:
\begin{center}
  \url{https://github.com/yanncalec/rundle}
\end{center}

\paragraph*{Acknowledgement}
The authors declare no competing financial interests or personal relationships that could influence the work reported in this paper. During manuscript preparation, large language models (LLMs) were used solely for text refinement and grammatical improvements. All technical content, theoretical developments, and experimental results remain the original contributions of the authors.

\newpage
\bibliography{STRF-Refs}

\newpage
\appendix
\renewcommand{\theequation}{\thesection.\arabic{equation}}
\section{Evidential Framework for Parameter Estimation}
\label{sec:ev_framework}
The \emph{evidence} function, or the marginalized likelihood, is the entry point to a principled way for Bayesian parameter estimation and uncertainty quantification, called the \emph{evidential framework} \citep{mackay_bayesian_1992}.

The classical MLE/MAP works at the \emph{likelihood level}, and estimates the likelihood parameters $\lambda$ with fixed hyperparameters $\xi$, via maximum likelihood $\ln\Pdf[\xi]{Y|\lambda}$ or maximum a posteriori $\ln \Pdf[\xi]{\lambda|Y}$. This only provides point estimators with frequentist confidence intervals. Moreover, the choice of hyperparameters $\xi$, which provides regularization and affects the model complexity through the prior, remains an issue.

The same MLE/MAP permits to extract more information by working at the \emph{evidential level}, also known as \emph{type-II MLE}. The hyperparameters $\xi$ are estimated by integrating out $\lambda$ and maximizing
\begin{align}
  \label{eq:LnEV_appendix}
  \ln \Pdf{\xi|Y} \simeq \ln \Pdf{Y|\xi} + \ln \pi\paren{\xi}
\end{align}
where $\pi(\cdot)$ is a hyperprior for $\xi$ that can be made uninformative. The evidence term $\Pdf{Y|\xi}$ here is given \eg by \eqref{eq:blr_evidence_case1} or \eqref{eq:blr_evidence_case2}.
From the estimated hyperparameters one obtains the posterior distribution $\lambda|Y$, and uncertainty of the predictive $Y_*|Y$ can be quantified through the closed-form posterior marginalization or Monte-Carlo sampling.

The evidential framework can also be used for optimization of high level parameters specific to a model $\lM$ (\eg polynomial degrees, interpolation kernel locations, or DNN architectures). At the \emph{model level}, we maximize $\ln\Pdf{Y|\lM}$ wrt model parameters by integrating out hyperparameters of the evidential level. This term is generally evaluated using the Laplace approximation:
\begin{align}
  \Pdf{Y|\lM} = \integral{\Pdf{Y|\xi, \lM}\cdot\pi(\xi)}{\xi} \approx \Pdf{Y|\ximap, \lM} \cdot \pi(\ximap) \cdot \paren{2\pi}^{\frac {\#\xi} 2}\abs{H}^{-\half}
\end{align}
where $\ximap$ is the optimal solution maximizing \eqref{eq:LnEV_appendix} under the model $\lM$, $\#\xi$ denotes the number of variables in $\xi$, and $H\defeq-\nabla^2_\xi\ln\Pdf{\ximap|Y, \lM}$ is the negative hessian matrix evaluated at $\ximap$.

The {evidential framework} solves first $\ximap$ at the evidential level, then use it for the estimation of likelihood parameters and for the model selection. This method has been shown to apply \emph{Occam's razor} principle: among all models that explain the observations, the one with the lowest complexity should be prefered.

\section{Some Convex Analysis} \label{sec:annex_convex}
Let $\Spp^n$ be the convex cone of $n$-square real symmetric and positive definite matrix. $\Spp^n$ can be equipped with the \emph{Löwner order}: for $A,B\in \Spp^n$, $A\preceq B$ if $B-A$ is positive semidefinite. A map $F:\Spp^n\to\Spp^m$ is \emph{monotone} if $A\preceq B$ implies $F(A)\preceq F(B)$. For a function $f:(0,\infty)\to (0,\infty)$, its action on $A\in\Spp^n$ is defined using the eigen value decomposition as
\begin{align}
  f[A] = U \diag(f(\lambda_i)) U^\top
\end{align}
with $U$ the eigen matrix and $(\lambda_i)_i$ the eigen values of A. Such a function is called \emph{operator-monotone} if for every order $n$ and any $A,B\in\Spp^n$ such that $A\preceq B$ it holds $f[A]\preceq f[B]$. Examples of operator-monotone function include $t\mapsto \ln(t)$ and $t\mapsto t^p$ with $0<p\leq 1$. An essential result from \citep{ando_concavity_1979} says that for an operator-monotone function $f$, the map $A\mapsto f[A]$ is concave on $\Spp^n$.

\begin{proposition}
  \label{prop:logdet_traceinv}
  The following statements hold:
  \begin{itemize}
    \item The log-determinant map $A\mapsto \ln\abs A$ is concave on $\Spp^n$ and (strictly) increasing, that is, $\ln\abs{A} \leq \ln\abs{B}$ if $A\preceq B$.
    \item For any $B\in\Spp^n$, the map $A\mapsto \tracep{\inv A B}$ is convex on $\Spp^n$ and (strictly) decreasing.
  \end{itemize}
\end{proposition}

The following well-known result has been used several times in our work. We include a proof here for the sake completeness.
\begin{corollary}
  \label{coro:logdet_traceinv_minimizer}
  For a $n$-square matrix $B$ which is symmetric and positive semidefinite, let $f$ be the real valued function defined on $\Spp^n$ as
  \begin{align}
    \label{eq:logdet_traceinv_def}
    f: A\mapsto \ln\abs{A} + \tracep{\inv A B}
  \end{align}
  Then $f$ admits $B$ as the unique minimizer on $\Spp^n$ if and only if $B$ is non-singular.
\end{corollary}
\begin{proof}
  First suppose $B\in\Spp^n$. By Proposition~\ref{prop:logdet_traceinv}, the map $g: A\mapsto f(\inv A)=-\ln\abs{A}+\tracep{AB}$ is convex, and matrix calculus gives $\partial_A g = -\inv A +B$. Hence applying the first order optimality condition $\partial_A g=0$ yields $\inv B$ as the unique minimizer of $g$, consequently $B$ is the unique minimizer of $f$.

  On the other hand, suppose $B$ has the eigen decomposition $B=U\diag(\lambda_i)U^\top$ with some $\lambda_i$ being 0. Then one can choose $A=U\diag(a_i)U^\top$ with $a_i>0$. In the limit that $a_i\to\lambda_i$ for every $i$, the value of $f(A)$ tends to $-\infty$.
\end{proof}

\section{Variational Posterior and BLL}
\label{sec:var_post_vbll}

We conduct Bayesian analysis for the same regression model of Section~\ref{sec:mvblr}, under the general setting that the baseline noise follows $\varepsilon~\sim\MvNormal{0}{U}$ where $U$ is unrelated to $V$, covariance of the prior $A\sim\MaNormal{M}{V}{K}$.


\begin{theorem}[Bayesian regression]
  \label{thm:blr_UV_case1}
  Let $E=Y-M\Phi$ and
  \begin{align}
    S \defeq D\otimes U + \paren{\Phit K \Phi}\otimes V, \ \ B \defeq \paren{\Phit K}\otimes V
  \end{align}
  where $\otimes$ denotes the Kronecker product. Under the same assumption of Theorem~\ref{thm:blr_case1} and with the same symbols, it holds the following distributions:
  \begin{itemize}
    \item The joint distribution is
    \begin{align}
      \vectorize{[A\ Y]} \sim \MvNormal{
        \begin{bmatrix}
          \vectorize{M}\\
          \vectorize{M\Phi}
        \end{bmatrix}
      }
      {
        \begin{bmatrix}
        K\otimes V & B^\top\\
        B & S
        \end{bmatrix}
      }
    \end{align}
    \item The evidence is
  \begin{align}
    \vectorize{Y} \sim \MvNormal{\vectorize{M\Phi}}{S}
  \end{align}
  \item The posterior is
  \begin{align}
    \vectorize{A}|Y \sim\MvNormal{\vectorize{M}+B^\top \inv S \vectorize{E}}{K\otimes V - B^\top \inv S B}
  \end{align}
  \item Given new inputs $X_*$, the posterior predictive is
  \begin{align}
    \vectorize{Y_*} | &Y \sim \lN_{\mathrm{vec}}  \left (
    \vectorize{M\Phi_*}+\paren{\Phit_* K \Phi} \otimes V \cdot \inv S \vectorize{E}, \right .\\
    & \left . D_*\otimes U + \paren{\Phit_* K \Phi_*}\otimes V - \paren{\Phit_* K \Phi}\otimes V \cdot \inv S \cdot \paren{\Phit K \Phi_*} \otimes V \right )
  \end{align}
  \end{itemize}
\end{theorem}
\begin{proof}[Sketch of proof]
  The proof is similar to that of Theorem~\ref{thm:blr_case1} and uses the fact a matrix-Normal distribution can be interpreted as a vector-Normal one.
\end{proof}

We can use a matrix-Normal variational posterior to approximate the true posterior which is too cumbersome to manipulate. The next result gives closed-form expression of Kullback-Leibler divergence between two matrix-Normal distributions. The proof is elementary and is left to the reader.
\begin{lemma}
  \label{lem:KL_normal}
  Let $\theta=\set{M,V,K}$ be parameters of the matrix-Normal distribution, and similarly for $\ttheta=\set{\tM,\tV,\tK}$. The Kullback-Leibler divergence between two distributions has the expression:
  \begin{align}
    \KL{\lN_\ttheta}{\lN_\theta} &= \frac\dmt 2 \ln\abs{V} +  \frac\dmy 2 \ln\abs{K} + \half \tracep{\inv V\tV}\tracep{\inv K \tK} + \\
    &\half\tracep{\inv V\paren{M-\tM}\inv K\trpp{M-\tM}} -\frac{\dmt}{2}\ln\abs{\tV} - \frac{\dmy}{2}\ln\abs{\tK}
  \end{align}
\end{lemma}

Now we can obtain the expression of ELBO objective \eqref{eq:elbo_def}, for the matrix-Normal prior $\lN_\theta$ and variational posterior $\lN_\ttheta$ with respectively parameters $\theta=\set{M,V,K}$ and $\ttheta=\set{\tM,\tV,\tK}$:
\begin{align}
  \label{eq:elbo_full_exp}
  & \ELBO{Y}{\lN_{\tilde\theta}}{\lN_\theta}
  \simeq - \frac{\dmy}{2}\ln\abs{D} - \frac{\dmt}{2}\ln\abs{V} - \frac{\dmy}{2}\ln\abs{K} + \frac{\dmt}{2}\ln\abs{\tV} + \frac{\dmy}{2}\ln\abs{\tK}\\
  & -\half\tracep{\inv U\paren{Y-\tM\Phi}\inv D\trpp{Y-\tM\Phi}} - \half \tracep{\Phi\inv D\Phit\tK} \tracep{\inv U\tV} \\
  & -\half\tracep{\inv V\tV}\tracep{\inv K\tK} - \half\tracep{\inv V\paren{M-\tM}\inv K \trpp{M-\tM}}
\end{align}

Notice that the role of ${\tilde\theta}$ and ${\theta}$ above are not symmetric. We consider the ELBO-maximization problem wrt the variational posterior:
\begin{align}
  \label{eq:elbo_post}
  \max_{\ttheta} \ELBO{Y}{\lN_\ttheta}{\lN_\theta}
\end{align}
and wrt the prior:
\begin{align}
  \label{eq:elbo_prior}
  \max_{\theta} \ELBO{Y}{\lN_\ttheta}{\lN_\theta}
\end{align}

\begin{proposition}\label{prop:elbo_theta_ttheta_specified}
  Let $V$ be fixed and shared in both the prior $\lN_\theta=\MaNormal{M}{V}{K}$ and the variational posterior $\lN_\ttheta=\MaNormal{\tM}{V}{\tK}$. Then
  \begin{itemize}
    \item For $\theta=\set{M,K}$ fixed and in the special case of $U=V$, the optimal posterior parameters $\ttheta=\set{\tM, \tK}$ maximizing \eqref{eq:elbo_post} are given by
    \begin{align}
      \label{eq:ttheta_specified}
      \tM=\SyxiSxx, \ \tK=\iSxx
    \end{align}
    \item For $\ttheta=\set{\tM,\tK}$ fixed, the optimal prior parameters $\theta=\set{M,K}$ maximizing \eqref{eq:elbo_prior} are given by
    \begin{align}
      \label{eq:theta_specified}
      M=\tM, \ K=\tK
    \end{align}
  \end{itemize}
\end{proposition}

\subsection{Equivalence of VBLL and one EM step}
\label{sec:equiv_vbll_em}

We have shown in Section~\ref{sec:EM_algo_case1} that for unregularized evidence $\ln\Pdf{Y|\xi}$, one EM step amounts to solve \eqref{eq:elbo_prior} for the optimal prior with the fixed variational posterior $\MaNormal{\tSyxiSxx}{V}{\inv\tSxx}$. By \eqref{eq:theta_specified} this corresponds to the joint update scheme
\begin{align}
  \label{eq:elbo_em_scheme_MK_flaw}
  M = \tSyx\inv\tSxx, \ \ K = \inv\tSxx.
\end{align}
On the other hand, variational BLL (VBLL) method \citep{harrison_variational_2024} seeks for the optimal variational posterior by solving \eqref{eq:elbo_post} with the fixed prior $\MaNormal{M}{V}{K}$, and reduces to \eqref{eq:ttheta_specified} in case of our BLL model. Clearly, the update \eqref{eq:ttheta_specified} and \eqref{eq:elbo_em_scheme_MK_flaw}, when applied iteratively from a same initialization $\paren{M_0, K_0}$, define a same sequence $\set{\paren{M_n, K_n}}$. The next result proves that the iteration \eqref{eq:elbo_em_scheme_MK_flaw} leads to the degenerate MLE of Theorem~\ref{thm:bll_mle_KVM}.

\begin{proposition}[Convergence of EM for unregularized evidence]
  \label{prop:conv_joint_scheme_unreg_ev}
  Assume that $\dmN\geq \max(\dmt,\dmy)$ and $\Phi$ has full rank, and all parameters except $M,K$ are fixed. The joint update scheme \eqref{eq:elbo_em_scheme_MK_flaw} converges to the degenerate MLE \eqref{eq:bll_mle_MK_case1} whatever the initialization for $M$ and $K$.
\end{proposition}
\begin{proof}
  By definition $\Sxx = \inv K + \Phi\inv D\Phit$ and applying the update $K=\inv\tSxx$ we see that at the $n$-th iteration
  \begin{align}
    \inv{K_n} = \inv{K_{n-1}} + \Phi\inv D\Phit = \inv{K_{0}} + n \Phi\inv D\Phit
  \end{align}
  therefore
  \begin{align}
    K_n \sim \inv n\invp{\Phi\inv D\Phit}
  \end{align}
  whatever the initialization $K_0$ for $K$. On the other hand, by definition of $\Syx$ in Section~\ref{sec:bayesian_regression}
  \begin{align}
    \tSyx\inv\tSxx 
    = Y \inv D\Phit\inv{\tSxx} + \tM \inv\tK\inv\tSxx
  \end{align}
  thus injecting the MLE solution $M$ in \eqref{eq:bll_mle_MK_case1} gives
  \begin{align}
    M - \hat M & =  Y \inv D\Phit\paren{\inv{\tSxx} - \invp{\Phi\inv D\Phit}} + \tM \inv\tK\inv\tSxx \\
    & = \paren{\tM - Y\inv D \Phit\invp{\Phi\inv D\Phit}} \inv\tK\inv\tSxx =  \paren{\tM - \hat M} \inv{\tK} K
  \end{align}
  and the $n$-th update $M_n$ satisfies
  \begin{align}
    M_n - \hat M = \paren{M_{n-1}-\hat M} \inv{K_{n-1}} K_n = \paren{M_0 - \hat M} \inv{K_0} K_n
  \end{align}
  Therefore $M_n\to \hat M$ whatever the initialization $M_0$ for $M$.
\end{proof}

\begin{figure}
  \centering
  \includegraphics[width=1.\textwidth]{{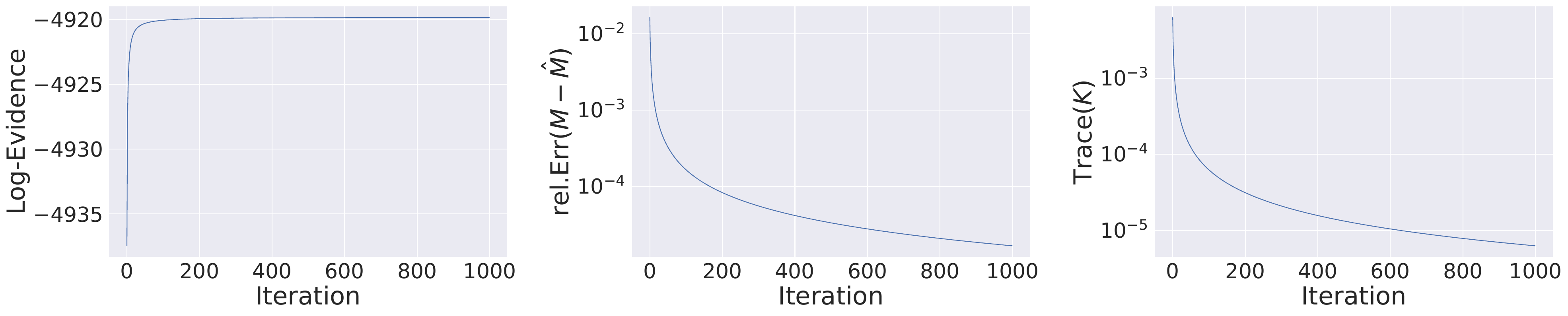}}
  \includegraphics[width=1.\textwidth]{{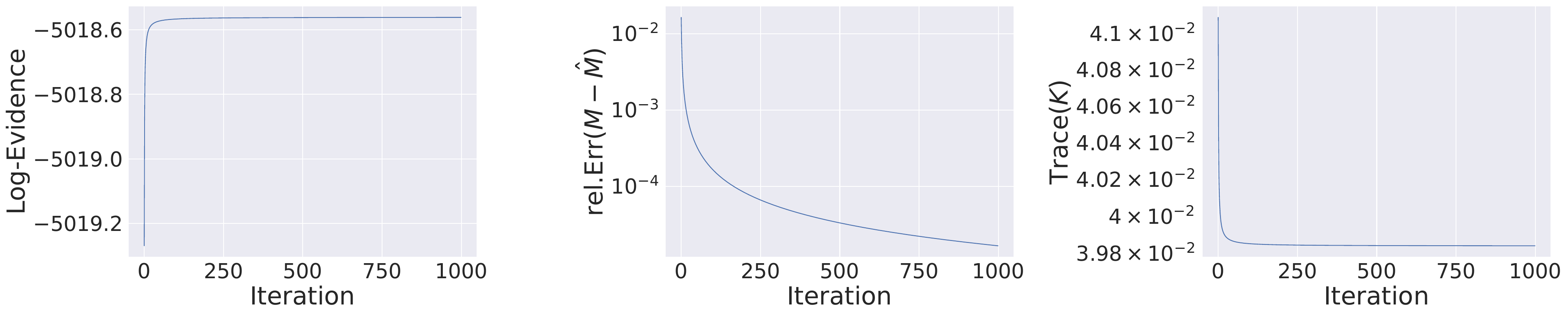}}
  \caption{Convergence of the EM algorithm using simulated data with random initialization. Top row: unregularized evidence using the update~\eqref{eq:elbo_em_scheme_MK_flaw}.  Bottom row: MAP with hyperprior on $K$ using the update~\eqref{eq:elbo_em_scheme_MK_fix_hyp}. Columns from left to right are respectively log-evidence, $\norm{M-\hat{M}}/\norm{\hat{M}}$, and $\trace(K)$.}
  \label{fig:flawed_em}
\end{figure}

Figure~\ref{fig:flawed_em} illustrates an example of convergence using simulated data.
For both unregularized (top) and hyperprior-regularized (bottom) evidence function, the log-evidence increases monotonically. It can be observed that $K_n \to 0$ in the unregularized case, as predicted by Proposition~\ref{prop:conv_joint_scheme_unreg_ev}. While an Inverse Wishart hyperprior on $K$ effectively prevents convergence to the degenerate solution.

\section{Proofs and Technical Details}\label{sec:proofs_tech}

\subsection{Theorem~\ref{thm:blr_case1}}
\begin{proof}
    The prior of $A$ and the likelihood $Y|A$ determines the pdf of the joint $[A\ Y]$:
    \begin{align*}
        \Pdf{A} \Pdf{Y|A} &\propto \exptrp{-\half \inv V \paren{\paren{A-M}\inv K \trpp{A-M} + \paren{Y-A\Phi}\inv D\trpp{Y-A\Phi} }} \\
        &=\exptrp{-\half\inv V \bracket{(A-M)\ \ (Y-M\Phi)}
        \begin{bmatrix}
            \inv K + \Phi\inv D\Phit & - \Phi\inv D\\
            -\inv D\Phit & \inv D
        \end{bmatrix}
        \begin{bmatrix}
            \trpp{A-M}\\\trpp{Y-M\Phi}
        \end{bmatrix}
        }
    \end{align*}
    Applying the inversion formula for block matrix, we identify from the expression above the pdf of the matrix-Normal distribution \eqref{eq:blr_joint_case1}. The other statements then follow from \eqref{eq:blr_joint_case1} by applying the conditionals of matrix-Normal distribution (see Theorem 2.3.12 of \citep{gupta_matrix_1999}).
\end{proof}

\subsection{Proposition~\ref{prop:bayes_leastsq_comparison}}
\begin{proof}
  Replacing $D$ by $\Omega$ in the least square problem \eqref{eq:lsq_solution_AD} defines another solution denoted by $\hat A_\Pxx$, which is actually identical to $\hat A_D$ as revealed by the identity in \eqref{eq:blr_helper_id_2}:
  \begin{align}
    \hat A_{D} = Y\inv D \Phit \invp{\Phi\inv D \Phit} =
    Y\iPxx \Phit \invp{\Phi\iPxx \Phit} = \hat A_{\Pxx}.
  \end{align}
  The right hand multiplier to $Y$ in $\hat A_\Pxx$ having full rank, the distribution of $\hat A_\Pxx$ is then determined from that of $Y$ as (Theorem 2.3.10 of \citep{gupta_matrix_1999}):
  \begin{align}
    \hat A_{\Pxx} \sim \MaNormal{M}{V}{\invp{\Phi \iPxx \Phit}}
  \end{align}
  Consequently the distribution of the least square estimator $\hat A_D$ follows from the helper identity in \eqref{eq:blr_helper_id_2} as
  \begin{align}
    \hat A_{D} \sim \MaNormal{M}{V}{K + \invp{\Phi\inv D \Phit}},
  \end{align}
  By using the helper identities in \eqref{eq:blr_helper_id_1}, the posterior mean $\Exp{A|Y} = \SyxiSxx$ follows the distribution
  \begin{align}
    \SyxiSxx\sim\MaNormal{M}{V}{K - \iSxx}.
  \end{align}
  The difference of the two variances then follows from the identity \eqref{eq:mvd_Normal_cov}.
\end{proof}

\subsection{Proposition~\ref{prop:blr_UQ_case1}}
\begin{proof}
  Under the assumption of sample independence, the distribution of $Y_*$ conditioned on $Y,\lambda,\theta$ is the same as conditioned on $\lambda$ alone. Therefore
  \begin{align*}
    \Pdf{Y_*|Y,\theta} = \integral{\Pdf{Y_*|\lambda}\Pdf{\lambda|Y,\theta}}{\lambda}
  \end{align*}
  which leads to
  \begin{align*}
    \Var{Y_*|Y, \theta} 
    &= \Exp{\Exp{\paren{Y_*-\Exp{Y_*|Y,\theta}}\trpp{Y_*-\Exp{Y_*|Y,\theta}}|\lambda}|Y,\theta}
  \end{align*}
  Writing $Y_*-\Exp{Y_*|Y,\theta}=\paren{Y_*-\Exp{Y_*|\lambda}}+\paren{\Exp{Y_*|\lambda}-\Exp{Y_*|Y,\theta}}$, the cross terms vanish since
  \begin{align*}
    \Exp{\paren{Y_*-\Exp{Y_*|\lambda}}\trpp{\Exp{Y_*|\lambda}-\Exp{Y_*|Y,\theta}}|\lambda} = 0,
  \end{align*}
  by consequence
  \begin{align*}
    \Var{Y_*|Y, \theta} &= \Exp{\Exp{\paren{Y_*-\Exp{Y_*|\lambda}}\trpp{Y_*-\Exp{Y_*|\lambda}}|\lambda}|Y,\theta} + \\
    & \Exp{\paren{\Exp{Y_*|\lambda}-\Exp{Y_*|Y,\theta}} \trpp{\Exp{Y_*|\lambda}-\Exp{Y_*|Y,\theta}}|Y, \theta}
  \end{align*}
  where the two terms of RHS correspond respectively to the aleatoric and the epistemic uncertainty in \eqref{eq:blr_total_var_decomp_predictive}, using the fact $\Exp{\Exp{Y_*|\lambda}|Y,\theta} =\Exp{Y_*|Y,\theta}$.

  The identity in \eqref{eq:mvd_Normal_cov} allows us to compute the covariance for matrix-Normal distribution. Given the likelihood
  \begin{align*}
    Y_*|\lambda\sim\MaNormal{A\Phi_*}{V}{D_*}
  \end{align*}
  we have $\Exp{Y_*|\lambda}=A\Phi_*$ and $\Var{Y_*|\lambda} = \trace\paren{D_*}V$. The aleatoric uncertainty \eqref{eq:blr_uncertainties_case1_aleatoric} then follows since $V$ is known. Since $A|Y,\theta \sim\MaNormal{\Syx\iSxx}{V}{\iSxx}$, therefore
  \begin{align*}
    \Var{A\Phi_*|Y,\theta} = \Exp{\paren{A-\Syx\iSxx}\Phi_*\Phi_*^\top\trpp{A-\Syx\iSxx}}
  \end{align*}
  and the epistemic uncertainty \eqref{eq:blr_uncertainties_case1_epistemic} follows by applying \eqref{eq:mvd_Normal_cov} again.
\end{proof}

As an example, Figure~\ref{fig:em_fixedbasis} depicts the decomposition of uncertainties computed using the formulae~\eqref{eq:blr_uncertainties_case1}.

\begin{figure}
  \centering
  \includegraphics[width=1.\textwidth]{{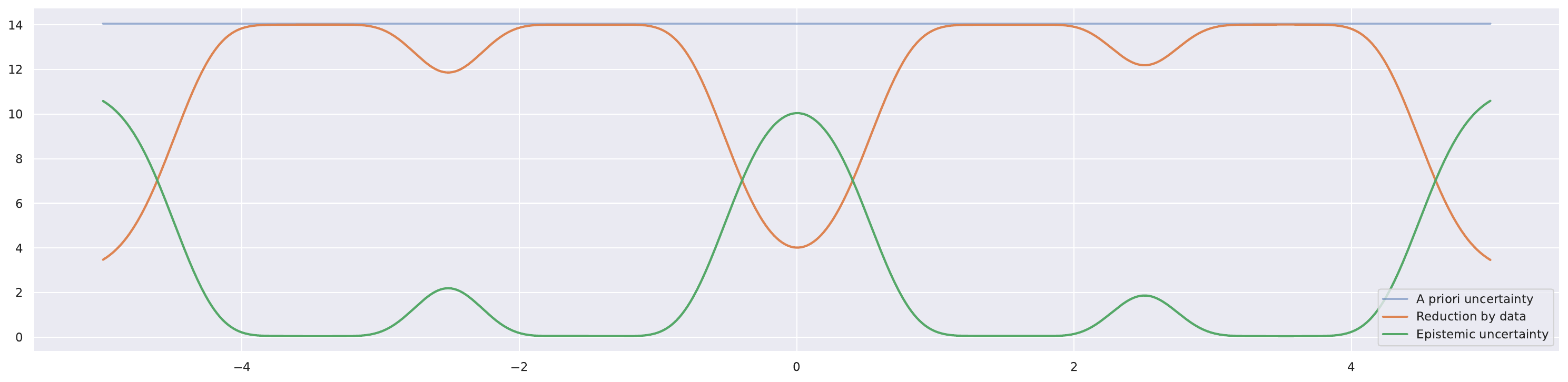}}
  \caption{Example of decomposition of epistemic uncertainty with the formula \eqref{eq:blr_case1_epistemic_decomp}.}
  \label{fig:decomp_epistemic}
\end{figure}

\paragraph*{An interpretation of epistemic uncertainty}
Using the expression of $\iSxx$ in \eqref{eq:blr_helper_id_1}, we can decompose the epistemic uncertainty \eqref{eq:blr_uncertainties_case1_epistemic} as
\begin{align}
  \label{eq:blr_case1_epistemic_decomp}
  \Var{A\Phi_*|Y} = \tracep{\Phi_*^\top K \Phi_*} V - \tracep{\Phi_*^\top K \Phi \iPxx \Phit K \Phi_*} V
\end{align}
This can be understood from the conditional distribution of $A\Phi_*|Y$. Actually, the joint distribution of $[A\Phi_*\ Y]$ is obtained from that of $[A\ Y]$ and reads
\begin{align}
  [A\Phi_*\ Y] \sim \MaNormal{[M\Phi_* \ M\Phi]}{V}{
    \begin{bmatrix}
      \Phi_*^\top K \Phi_* & \Phi_*^\top K \Phi\\
      \Phit K \Phi_* & \Pxx
    \end{bmatrix}
  }
\end{align}
So the first term in \eqref{eq:blr_case1_epistemic_decomp} corresponds to the covariance $\Var{A\Phi_*|\theta}$ that can be interpreted as the \apriori uncertainty of the prediction at $\Phi_*$. Due to the observation of $Y$, the true epistemic uncertainty is obtained from the \apriori uncertainty after variance reduction corresponding to the second term in \eqref{eq:blr_case1_epistemic_decomp}. An example of decomposition corresponding to Figure~\ref{fig:em_fixedbasis} is visualized in Figure~\ref{fig:decomp_epistemic}.

\subsection{Proposition~\ref{prop:asymptotic_epistemic}}
\begin{proof}
  Under the law of large number, the sample matrix $\Phi\inv D\Phit / \dmN$ converges to the left hand side of \eqref{eq:PDP_condition} almost surely, implying also the convergence of all eigen values. In particular the $k$-th eigen value of ${\Phi\inv D\Phit}$ scales as $\rho_k = O(\dmN)$. The inverse $\invp{\Phi\inv D\Phit}$ shares the same eigen vectors $\set{u_k}$ of $\Phi\inv D\Phit$ but its eigen values scale as $O(1/\dmN)$, thus $\invp{\Phi\inv D\Phit}$ converges to 0 in Frobenius norm. Actually, since $(u_k u_k^\top)_{ij}$ is bounded, we have
  \begin{align*}
    \invp{\Phi\inv D\Phit}_{ij} = \sum_{k=1}^\dmt \rho_k^{-1} (u_k u_k^\top)_{ij} = O(1/\dmN)
  \end{align*}
  As consequence, using the helper identity in  \eqref{eq:blr_helper_id_1} and \eqref{eq:blr_helper_id_2} it holds almost surely
  \begin{align*}
    \Phi\iPxx\Phit = \invp{K+\invp{\Phi\inv D\Phit}} = \inv K - \inv K O(1/\dmN) \inv K
  \end{align*}
  as well as
  \begin{align*}
    \iSxx = K - K \Phi\iPxx\Phit K = O(1/\dmN).
  \end{align*}
  as the number of samples increases.
\end{proof}

\subsection{Theorem~\ref{thm:bll_mle_KVM}}
\begin{proof}
  The loss is quadratic in $M$, therefore for $K$ fixed, setting $\partial_M\lL=0$ in \eqref{eq:bll_mle_derv_case1} yields the unique minimizer of $M$ which reads
  \begin{align}
    \hat M &= Y \iPxx{\Phi}^\top \invp{\Phi\iPxx{\Phi}^\top} =  Y {\inv D} \Phit \invp{\Phi{\inv D} \Phit}
  \end{align}
  where we applied the helper identity in \eqref{eq:blr_helper_id_2}. Since $Y\sim\MaNormal{M\Phi}{V}{\Pxx}$, the unbiasedness of $\hat M$ can be easily verified by taking the expectation.

  We define
  \begin{align}
    \label{eq:projector_P_PhiD}
    P\defeq I_\dmN - \pinvp{\Phi\hinv D}\Phi\hinv D
  \end{align}
  to be the orthogonal projector onto the kernel space of $\Phi\hinv D$ where $\dagger$ denotes the generalized inverse. With $\hat M$ as $M$, the evidential loss \eqref{eq:ev_loss_case1} is reduced to the following form
  \begin{align}
    \label{eq:MLE_M_Loss_form}
    \lL = \frac{\dmy}{2}\ln\abs\Pxx + \half\tracep{Y^\top\inv V Y \hinv D P \hinv D},
  \end{align}
  which contains $K$ only through $\ln\abs\Pxx$, which is concave and increasing in $\Pxx$ by Proposition~\ref{prop:logdet_traceinv}. We can prove that as function of $K$, $K\mapsto \ln\abs{\Pxx(K)}$ is strictly increasing. For this, let $W$ be such that ${W W^\top} = \iPxx$ and suppose there exists $H\in\Spp$ such that $\ln\abs{\Pxx(K)}=\ln\abs{\Pxx(K+H)}$. Then it is equivalent to
  \begin{align*}
    \ln\abs{I_\dmN + \trpp{\Phi W} H \Phi W} = 0
  \end{align*}
  which is in contradiction with the statement of Lemma~\ref{lem:bll_det}.
  Consequently for any $K\in\Spp$ it holds $\lL(K) > \lL(aK)$ as the constant $a\to 0^+$, implying $\lL(K)$ is minimized at $\hat K=0$.
\end{proof}

\begin{lemma}\label{lem:bll_det}
  Let $U$ be a full rank matrix. It holds for any $A\in\Spp$ of appropriate shape that $\ln\abs{I + U^\top A U}>0$.
\end{lemma}
\begin{proof}
  Suppose there exists $A\in\Spp$ for the contrary claim to hold, then the singular value decomposition shows that the matrix $U^\top A U$ must be identically zero.  This implies $0=\tracep{A U U^\top} \geq \lambda_* \tracep{U U^\top}$ where $\lambda_*>0$ is the smallest singular value of $A$. This is in contradiction with the fact of $U$ having full rank.
\end{proof}

\subsection{Theorem~\ref{thm:blr_case2}}

First we state an elementary result on the compound Normal-Inverse Wishart distribution:
\begin{proposition}
  \label{prop:mvd_Normal_InvWishart}
  Let $V\sim p\times p$ follow $\InvWishart{\Psi}{\Ndf}$, and $A\sim p\times n$ conditioned on $V$ follow $\MaNormal{M}{V}{K}$ for some fixed $K$. Then the joint distribution of $[V \ A]$ is 
  \begin{align*}
    [V \ A]\sim \MaNormal{M}{V}{K} \otimes \InvWishart{\Psi}{\Ndf}.
  \end{align*}
  The marginal distribution of $A$ is
  \begin{align*}
    A \sim \MaStudent{M}{\Psi}{K}{\Ndf-2p}.
  \end{align*}
  and the conditional of $V|A$ is
  \begin{align*}
    V|A \sim \InvWishart{\Psi+\paren{A-M}K^{-1}\paren{A-M}^\top}{\Ndf+n}
  \end{align*}
\end{proposition}

The proof of Theorem~\ref{thm:blr_case2} is given in the following.

\begin{proof}
  Various paths can lead to these same results and below is just one of them. The followings have the same distribution as their counterpart in the known $V$ case:
  \begin{enumerate}
    \item $[A\ Y]|V$ is distributed as \eqref{eq:blr_joint_case1};
    \item $A|V$ is distributed as \eqref{eq:blr_prior_case1};
    \item $Y|V$ is distributed as \eqref{eq:blr_evidence_case1}.
  \end{enumerate}
  All being matrix-Normal distributed, therefore applying Proposition~\ref{prop:mvd_Normal_InvWishart} on each gives
  \begin{enumerate}
    \item the joint marginal \eqref{eq:blr_joint_AY_case2}, as well as the posterior update \eqref{eq:blr_post_update_V_case2};
    \item the posterior marginal \eqref{eq:blr_marginal_A_case2};
    \item the evidence \eqref{eq:blr_evidence_case2} as well as the posterior marginal \eqref{eq:blr_post_marginal_V_case2}.
  \end{enumerate}
  Since $[A\ V]|Y$ is the product of $A|V,Y$ and $V|Y$, which are distributed respectively as \eqref{eq:blr_posterior_case1} and \eqref{eq:blr_post_marginal_V_case2}, its distribution is a compound Normal-Inverse Wishart. Therefore calling again Proposition~\ref*{prop:mvd_Normal_InvWishart} gives the marginal \eqref{eq:blr_post_marginal_A_case2}.
  Applying \eqref{eq:blr_evidence_case2} on $[Y_*\ Y]$ gives
  \begin{align*}
    [Y_*\ Y] \sim \MaStudent{[M\Phi_*\ M\Phi]}{\Spr}{
      \begin{bmatrix}
        D_*+\Phi_*^\top K \Phi_* & \Phi_*^\top K \Phi \\
        \Phit K \Phi_* & D + \Phit K \Phi
    \end{bmatrix}}{\Ndf-2\dmy}
  \end{align*}
  the posterior predictive \eqref{eq:blr_predictive_case2} then follows from the conditionals of matrix-T distribution, see Theorem 4.3.9 of \cite{gupta_matrix_1999}. Finally \eqref{eq:blr_predictive_case2_A} can be established in the same way by working on the joint $[A \ Y \ Y_*]$ using \eqref{eq:blr_joint_AY_case2}.
\end{proof}

\subsection{Proposition~\ref{prop:blr_UQ_case2}}
\begin{proof}
  Same as the proof of Proposition~\ref*{prop:blr_UQ_case1}, it holds $Y_*|\lambda\sim\MaNormal{A\Phi_*}{V}{D_*}$. Moreover, $V|Y,\theta$ and $A|Y,\theta$ are given respectively by \eqref{eq:blr_post_marginal_V_case2} and \eqref{eq:blr_post_marginal_A_case2}. The aleatoric uncertainty then follows from the mean of inverse Wishart distribution, and the epistemic uncertainty follows by applying \eqref{eq:mvd_T_cov}.
\end{proof}

\subsection{Theorem~\ref{thm:bll_mle_KVM_T}}
\begin{proof}
  Matrix calculus gives
  \begin{align}
    \pypx{\lL}{M} &= -(\Nprp+\dmN) \inv H E \iPxx \Phi^\top.
  \end{align}
  The first order optimality criteria $\partial_M\lL=0$ yields a critical point which coincides with $\hat M$ of \eqref{eq:bll_mle_MK_case1}, and is independent of $K, \Spr$. To establish that $\hat M$ is a global minimizer, we must verify that the hessian $\partial_M^2 \lL$ evaluated at $\hat M$ is positive. Let $U$ be a matrix with the same dimension as $M$, then, the quadratic form of $\partial_M^2\lL|_M$ applied to $U$ is
  \begin{align*}
    \seq{\partial_M^2\lL|_M(U), U} = -\frac{\Nprp+\dmN}{2} \tracep{\partial_M\tracep{\inv H E \iPxx \Phi^\top U^\top} U^\top}
  \end{align*}
  where the inner term can be obtained using matrix calculus as
  \begin{align*}
    \partial_M\tracep{\inv H E \iPxx \Phi^\top U^\top} = - & \inv H U \Phi \iPxx \Phi^\top + \\ & \inv H\paren{U\Phi\iPxx E^\top
    + E\iPxx\Phi^\top U^\top}\inv H E\iPxx\Phi^\top
  \end{align*}
  Notice that the last term evaluated at $\hat M$ vanishes. Actually, recall that
  \begin{align}
    \hat E = Y-\hat M \Phi = Y\hinv D P D^\half
  \end{align}
  and $P$ is the projector onto the kernel space of $\Phi\hinv D$, therefore
  \begin{align*}
    \hat E \iPxx \Phi^\top = Y \hinv D P \hinv D \Phi^\top - Y \hinv D P \hinv D \Phi^\top \iSxx \Phi \hinv D = 0.
  \end{align*}
  Since by construction $H$ is always positive definite, consequently,
  \begin{align*}
    \seq{\partial_M^2\lL|_{\hat M}(U), U} = \frac{\Nprp+\dmN}{2} \tracep{\inv{\hat H} U \Phi\iPxx\Phi^\top U^\top} \geq 0
  \end{align*}
  which proves that $\partial_M^2\lL$ evaluated at $\hat M$ is positive. This proves also that $\hat M$ is the global minimizer of $\lL$ since it is the unique critical point.

  Similarly, we have the derivative (without taking into account the symmetry of $\Spr$)
  \begin{align}
    \pypx{\lL}{\Spr} &= \half\paren{(\Nprp+\dmN) \inv H - \Nprp \inv\Spr}.
  \end{align}
  The solution $\hat \Spr$ in \eqref{eq:bll_mle_V_case2} is obtained from the first order optimality criteria $\partial_\Spr \lL=0$, which gives
  \begin{align*}
    \Spr = \frac{\Nprp}{\dmN} E\iPxx E^\top.
  \end{align*}
  Substituting $\hat M$ and after simplification, we get \eqref{eq:bll_mle_V_case2}.
  Applying the identity \eqref{eq:mvd_T_cov} with $Q\defeq \hinv D P \hinv D$ gives
  \begin{align*}
    \Exp{\hat\Spr} = \frac{\Nprp}{\dmN} \Exp{\paren{Y-M\Phi} Q \trpp{Y-M\Phi}} =  \frac{\Nprp}{\dmN} \cdot \frac{\dmN-\dmt}{\Npr-2\dmy-2} \Spr.
  \end{align*}
  To verify that $\hat \Spr$ of \eqref{eq:bll_mle_V_case2} is a global minimizer, we fix $\hat M$ and compute the Hessian $\partial_\Spr^2 \lL$ and check its positive definiteness at $\hat\Spr$.\footnote{The cross term $\partial_\Spr\partial_M\lL$ vanishes at $\hat M$ since $\hat M$ does not depend on $\Spr$. This is not true in general.} Let $U$ be a symmetric matrix of the same shape as $\Spr$, then the quadratic form of $\partial_\Spr^2 \lL$ applied on $U$ is
  \begin{align*}
    \seq{\partial_\Spr^2 \lL|_{\Spr}(U), U} &= \tracep{\partial_\Spr\tracep{\partial_\Spr\lL|_\Spr U}U} \\
    &= -\frac{\Nprp+\dmN}{2} \tracep{\inv H U \inv H U} + \frac{\Nprp}{2} \tracep{\inv \Spr U \inv \Spr U}.
  \end{align*}
  Substituting $\hat \Spr$ and $\hat M$, we get
  \begin{align*}
    \hat H = \frac{\Nprp+\dmN}{\dmN} Y \hinv D P \hinv D Y^\top = \frac{\Nprp+\dmN}{\Nprp} \hat\Spr
  \end{align*}
  which is invertible with probability 1. Now injecting $\hat H$ into the expression gives
  \begin{align*}
    \seq{\partial_\Spr^2 \lL|_{\hat\Spr}(U), U} = \frac{\dmN\Nprp}{2\paren{\Nprp+\dmN}}\tracep{\inv{\hat\Spr}U\inv{\hat\Spr}U}
  \end{align*}
  Therefore, $\seq{\partial_\Spr^2 \lL|_{\hat \Spr}(U), U} \geq 0$ for all $U$, proving that $\partial_\Spr^2 \lL$ is positive at $\hat \Spr$. Consequently, $\hat \Spr$ is a local minimizer of $\lL$ with respect to $\Spr$, hence the global minimizer since it is the unique critical point.

  Finally, with $\hat M$ and $\hat\Spr$ the only term of the loss function containing $K$ is $\ln\abs{\Omega}$, and the proof of the last statement is identical to the counterpart in Theorem~\ref{thm:bll_mle_KVM}.
\end{proof}

\section{Complementary Numerical Experiments}
\label{sec:annexe_numexp}
\def\figwidth{1.}


\subsection{A toy problem}
\begin{figure}[H]
  \centering
  \includegraphics[width=.9\textwidth]{{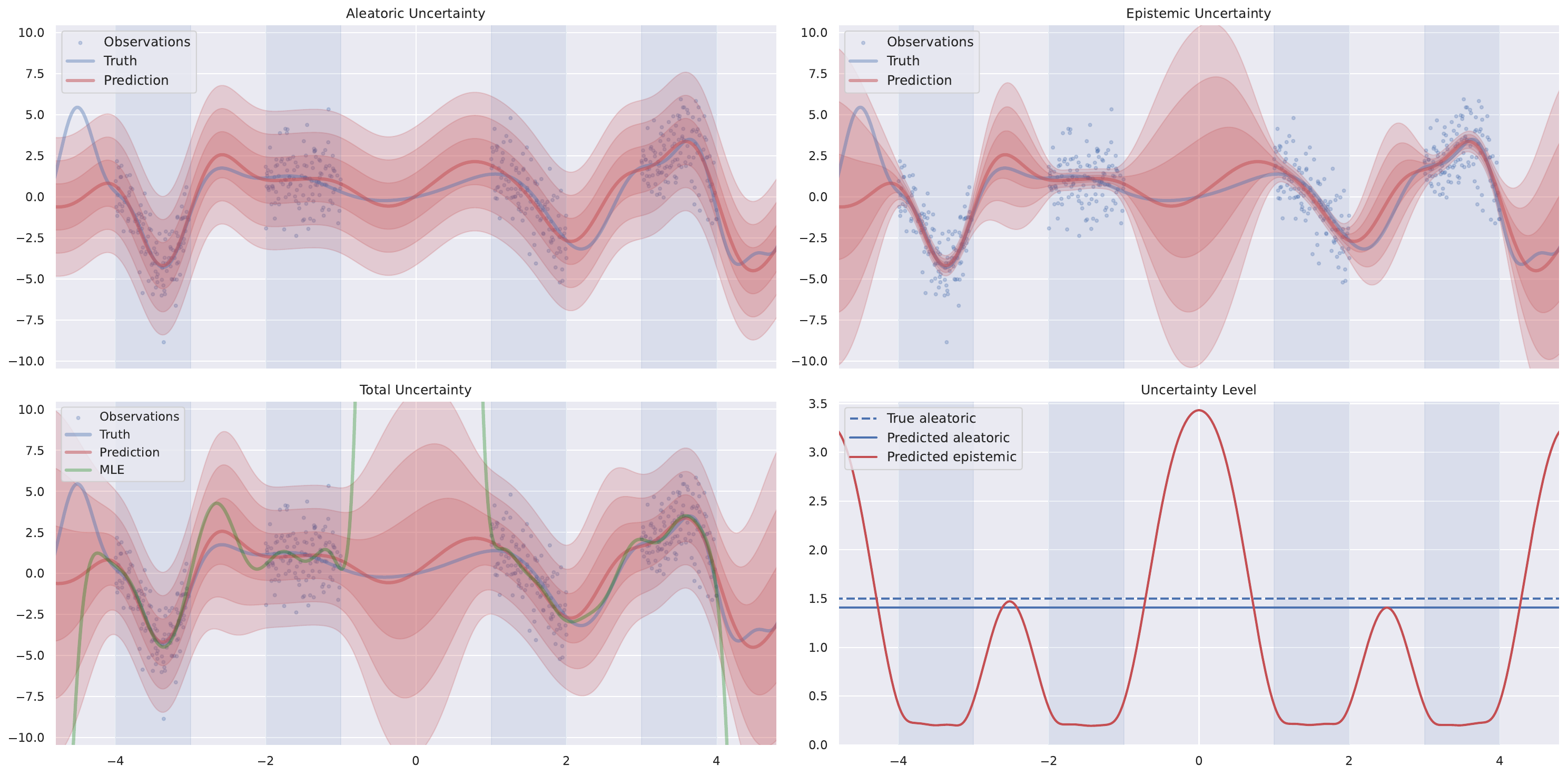}}
  \caption{Algorithm~\ref{alg:ELBO_EM_framework} for Bayesian interpolation with fixed basis (Gaussian kernels) and homoscedastic noise.}
  \label{fig:em_fixedbasis}
\end{figure}

We demonstrate the comparative performance of MLE (obtained via the least square minimization) and Bayesian predictors through a toy example, as shown in the bottom-left pannel of Figure~\ref{fig:em_fixedbasis}. The data collection protocol is similar to the experiment in Figure~\ref{fig:em_dnn} but we assume here a constant noise level. For Bayesian predictor we set $M=0$ and estimate the optimal isotropic $K=kI$ and constant $\sigma$ using the specified Algorithm~\ref{alg:ELBO_EM_Interpolation}.

\begin{algorithm}
  \caption{EM algorithm for Bayesian interpolation}\label{alg:ELBO_EM_Interpolation}
  \begin{algorithmic}
    \Require $X, Y, M, V$, initialization $\xi_0=\set{\sigma^2_0, K_0}$, maximum iteration $\Nmax$
    \State $n\leftarrow 0$
    \While{$n<\Nmax$}
    \State   $\tilde\xi \leftarrow \xi_{n-1}$, that is, $\tilde\sigma^2=\sigma_{n-1}^2, \tK = K_{n-1}$
    \State   (E-step) Formulate $Q_1$ with $\txi$
    \State   (M-step for $\sigma^2$) Maximization of $Q_1$ with respect to $\sigma^2$
    \begin{align}
      \sigma^2 = \frac{1}{\dmy\dmN}{\tracep{\tE^\top \inv V \tE + \dmy\Phit\inv\tSxx\Phi}}
    \end{align}
    \State Compute  $\tG = \inv\tSxx + \inv \dmy {\trp\tF\inv V \tF}$
    \State   (M-step for $K$) Depending on whether $K=\diag(k_j)$ or $K=k I_\dmt$, update
    \begin{align}
      k_j &= \tG_{jj}, \ \text{ or } k = \frac{1}{\dmt} \tracep{\tG}
    \end{align}
    \State $\xi_n\leftarrow$ the updated $\set{\sigma^2, K}$
    \State $n\leftarrow n+1$
    \EndWhile
  \end{algorithmic}
\end{algorithm}


\subsection{More numerical results}

\begin{figure}[htp]
  \centering
  \includegraphics[width=1.\textwidth]{{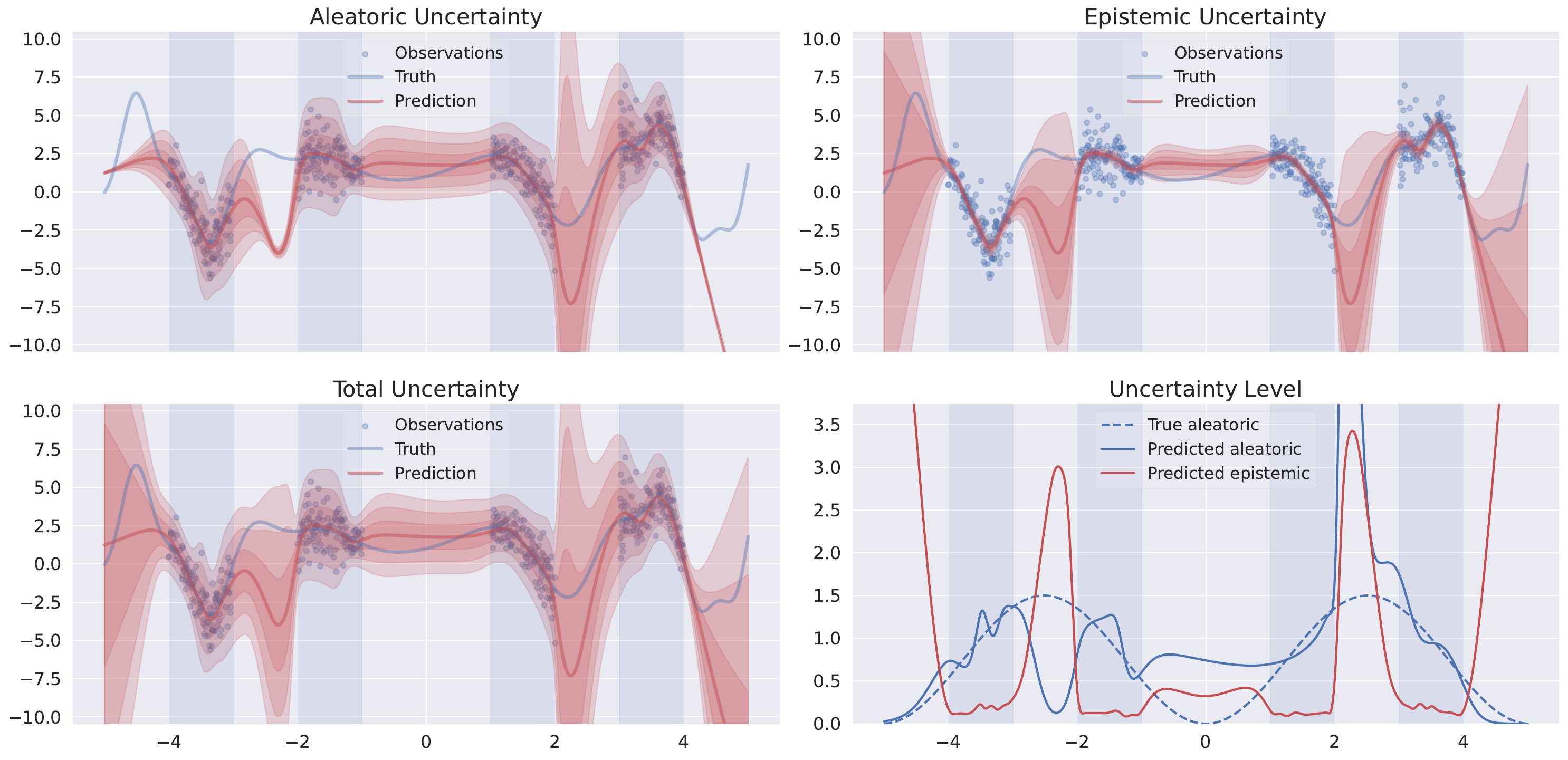}}
  \caption{Same experiment as Figure~\ref{fig:em_dnn} with different initialization and hyperprior parameters.}
  \label{fig:em_dnn2}
\end{figure}

\begin{figure}[htp]
  \centering
  \includegraphics[width=1.\textwidth]{{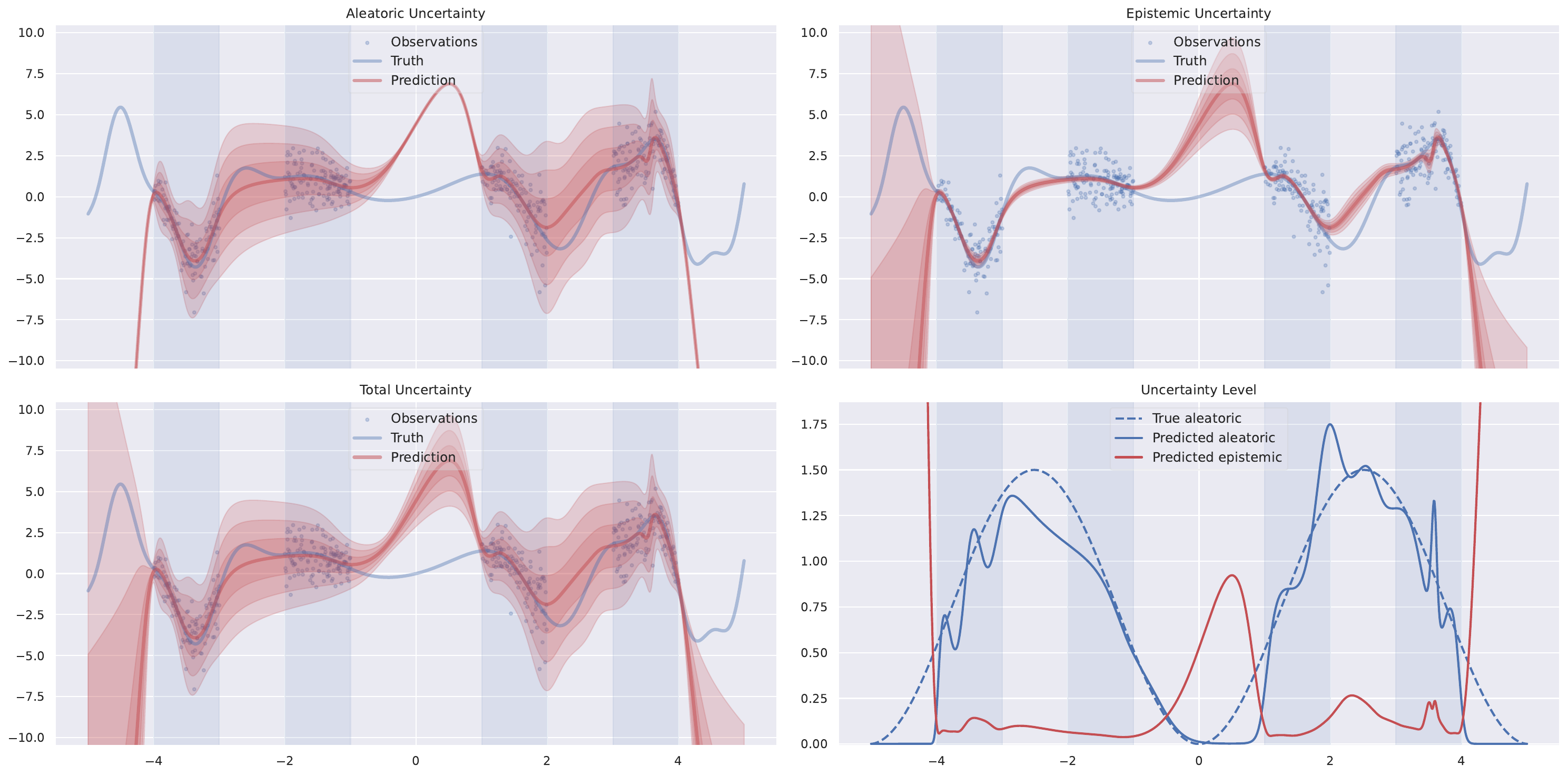}}
  \caption{Same experiment as Figure~\ref{fig:em_dnn} with 32 units per layer in the MLP.}
  \label{fig:em_dnn32}
\end{figure}


\begin{figure}[htp]
  \centering
  \includegraphics[width=1.\textwidth]{{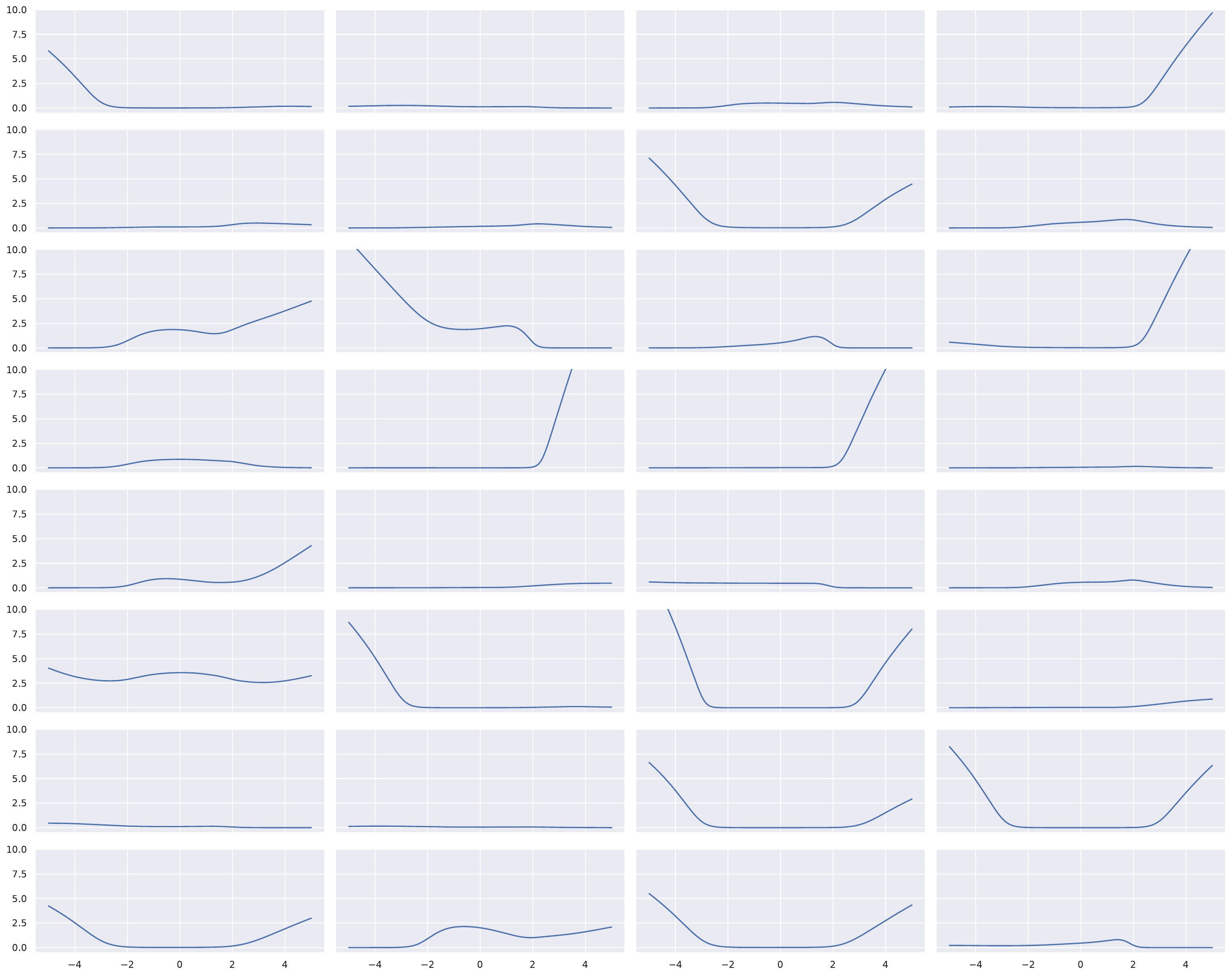}}
  \caption{Example of basis functions learned by a probability density network to fit the data in Figure~\ref{fig:em_fixedbasis}.}
  \label{fig:learned_basis}
\end{figure}

\begin{figure}
  \centering
  \includegraphics[width=1.\textwidth]{{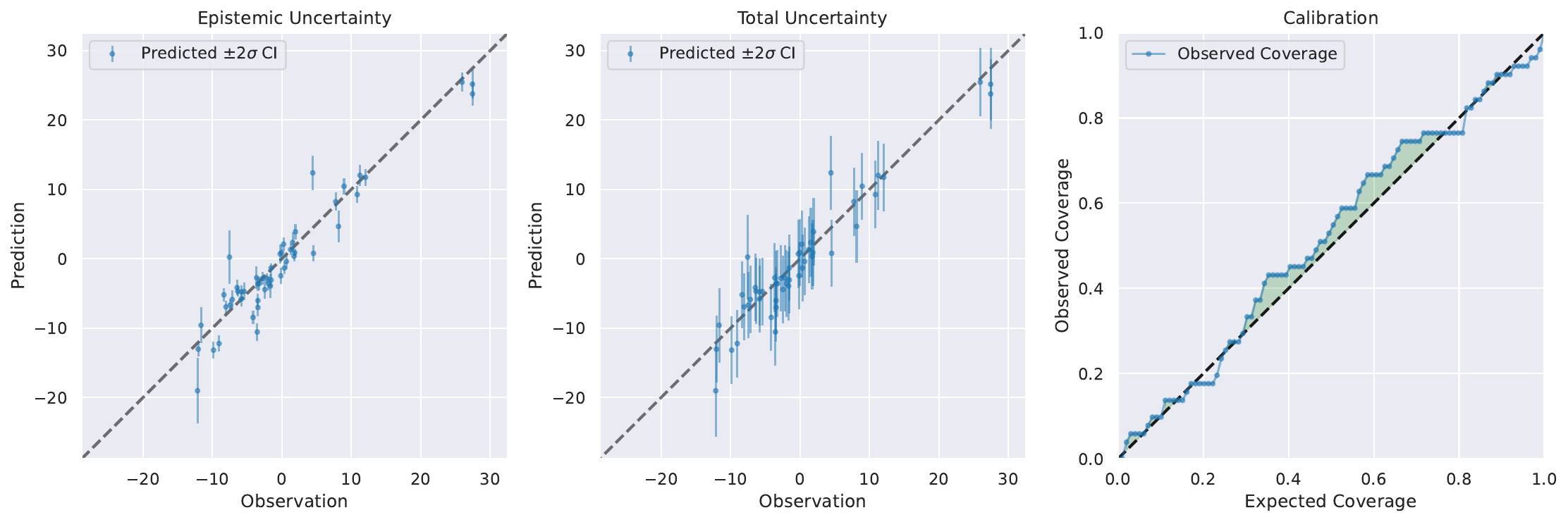}}
  \caption{An instance of uncertainties and calibration curve on the test data of dataset \textsc{Boston}. Predicted posterior means are plotted against observations  with $2\sigma$ epistemic (left) and total uncertainty (center). Calibration curve (right) is evaluated with the total uncertainty.}
  \label{fig:uq_cali_boston}
\end{figure}




\begin{figure}
  \centering
  \includegraphics[width=1.\linewidth]{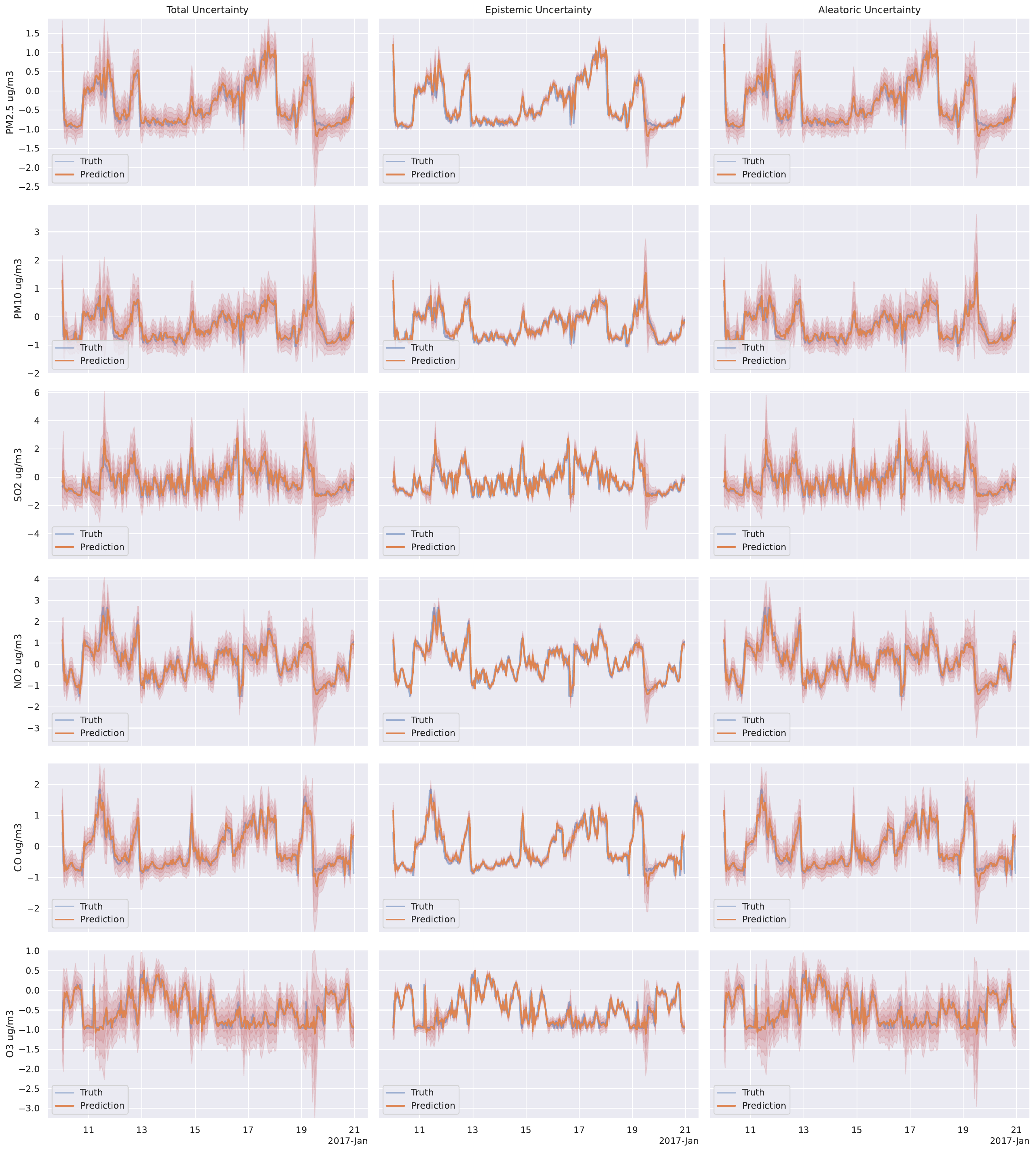}
  \caption{Same experiment as in Figure~\ref{fig:beijing} applied on the period Jan. 10 to Jan. 20, 2017, where the epistemic uncertainty returned to the baseline level, signaling the disappearance of the previous distribution shift in data space.}
  \label{fig:beijing_3}
\end{figure}

\end{document}